\documentclass[sigconf]{acmart}
\AtBeginDocument{%
  }

\usepackage{pifont}
\usepackage{booktabs} 
\usepackage{multirow} 
\usepackage{amsmath} 
\usepackage{xcolor} 
\usepackage{colortbl} 
\usepackage{graphicx} 
\usepackage{subcaption}
\usepackage{graphicx}
\usepackage{makecell}
\usepackage{adjustbox}
\usepackage{booktabs}
\usepackage{multirow}
\usepackage[table,xcdraw]{xcolor}
\usepackage[ruled,vlined]{algorithm2e}
\usepackage[most]{tcolorbox}
\usepackage{xcolor}

\definecolor{gcfusecolor}{HTML}{E6F0FF} 
\definecolor{foundationcolor}{HTML}{FFF4E6} 
\definecolor{lightgray}{HTML}{F2F2F2} 
\theoremstyle{definition}
\newtheorem{definition}{Definition}[section]

\definecolor{gcfusecolor}{gray}{0.9} 
\definecolor{gcfusecolor}{RGB}{230, 240, 255}

\definecolor{syntaxblue}{RGB}{0, 102, 204}
\definecolor{syntaxgray}{RGB}{120, 120, 120}

\newcommand{\gcfuseIcon}{%
    \protect\textsf{%
        \protect\textcolor{syntaxgray}{\textbf{\{}}%
        \protect\textcolor{syntaxblue}{\textbf{GS-Fuse}}%
        \protect\textcolor{syntaxgray}{\textbf{\}}}%
    }%
}

\setcopyright{acmlicensed}
\copyrightyear{2026}
\acmYear{2026}
\setcopyright{cc}
\setcctype{by}
\acmConference[KDD 2026]{Proceedings of the 32nd ACM SIGKDD Conference on Knowledge Discovery and Data Mining V.2}{August 9--13, 2026}{Jeju Island, Republic of Korea.}
\acmBooktitle{Proceedings of the 32nd ACM SIGKDD Conference on Knowledge Discovery and Data Mining V.2 (KDD '26), August 09--13, 2026, Jeju Island, Republic of Korea}
\acmDOI{10.1145/3770855.3817927}
\acmISBN{979-8-4007-2259-2/2026/08}

\begin{document}

\title[\gcfuseIcon: Granger-Gated Multimodal Fusion for Event-Driven Financial Forecasting]{\gcfuseIcon: Granger-Supervised Gated Fusion and Multi-Granularity Alignment for Event-Driven Financial Forecasting}

\author{Yang Zhang}
\authornote{Corresponding authors.}
\email{zhang.yang.r54@kyoto-u.jp}
\orcid{0000-0003-4261-3801}
\affiliation{%
  \institution{Southwestern University of Finance and Economics}
  \city{Chengdu}
  \country{China}
}
\author{En Chun}
\email{enchun33@gmail.com}
\affiliation{%
  \institution{Southwestern University of Finance and Economics}
  \city{Chengdu}
  \country{China}
}
\author{Ziyun Mao}
\email{224120100004@smail.swufe.edu.cn}
\affiliation{%
  \institution{Southwestern University of Finance and Economics}
  \city{Chengdu}
  \country{China}
}
\author{Yulu Wu}
\email{225120100001@smail.swufe.edu.cn}
\affiliation{%
  \institution{Southwestern University of Finance and Economics}
  \city{Chengdu}
  \country{China}
}
\author{Jun Wang}
\authornotemark[1]  
\email{wangjun1987@swufe.edu.cn}
\affiliation{%
  \institution{Southwestern University of Finance and Economics}
  \city{Chengdu}
  \country{China}
}

\renewcommand{\shortauthors}{\textbf{Yang Zhang} et al.}

\begin{abstract}
    Accurately forecasting the impact of salient financial events on markets is critical for investors and policymakers. However, existing multimodal time-series models typically fuse text and prices symmetrically, without an explicit way to decide when event text is truly predictive, and thus struggle to exploit the directional event-to-price structure and the heterogeneous roles of textual and price signals. In this work, we propose GS-Fuse, a multimodal event-based forecasting framework that employs (i) a Granger-supervised, causal-aware gated fusion module, which learns to open toward event text only when it provides incremental predictive value beyond historical prices, and (ii) a multi-granularity alignment mechanism that jointly aligns high-level event representations and fine-grained textual cues with future market trajectories. Built as a flexible, plug-and-play adapter on top of off-the-shelf large language models and time-series foundation models, GS-Fuse can be instantiated across diverse backbones and market settings. Extensive experiments on real-world financial datasets show that GS-Fuse consistently outperforms state-of-the-art time-series and multimodal baselines across multiple assets and forecasting horizons. The implementation is publicly available in our \href{https://github.com/lakebodhi/GS-FUSE}{\textcolor{blue}{\textbf{GitHub repository}}}.

\end{abstract}




\maketitle

\begin{figure*}
    \centering
    \includegraphics[width=0.9\linewidth]{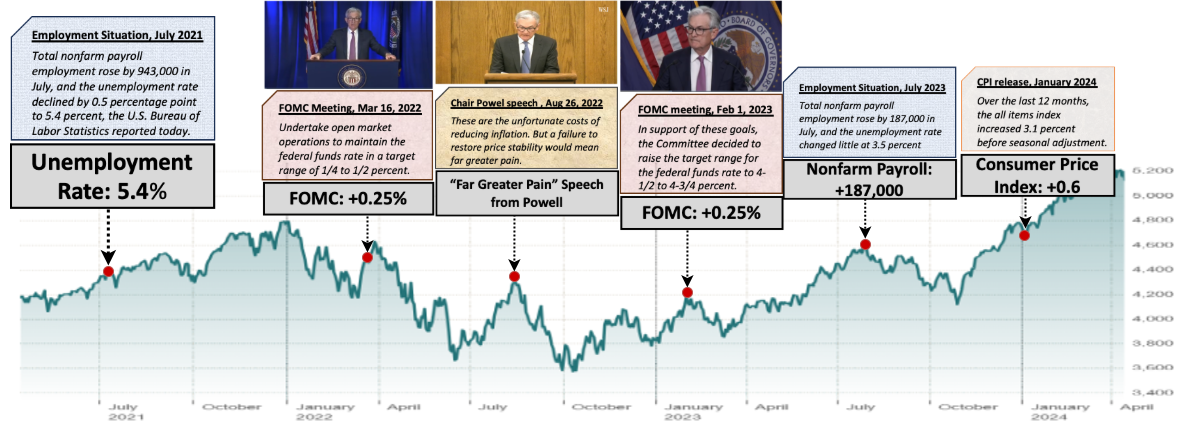}
    \caption{Illustration of financial event–based forecasting. The S\&P500 index exhibits pronounced movements around key macro and policy events: employment situation reports (unemployment rate 5.4\% in July 2021; nonfarm payroll +187{,}000 in July 2023), FOMC meetings with +0.25\% rate hikes (March 16, 2022 and February 1, 2023), Chair Powell's ``far greater pain'' speech (August 26, 2022), and a CPI release (+0.6 in January 2024).}
    \label{fig:event_demo}
\end{figure*}

\section{Introduction}
Fama's Efficient Market Hypothesis (EMH) posits that financial asset prices fully incorporate all available information and adjust when new information enters the market \cite{1731c543-0a2e-3dd1-85c2-3ffc09a485a7,78bd1d1d-3f88-3c53-ad25-c8ac73767958}. Event-based financial forecasting is a classic problem in finance, and extensive empirical work shows that major macroeconomic announcements, including monetary policy decisions, inflation releases, employment reports, and GDP revisions, can trigger significant market movements by reshaping investors' expectations about growth, risk, and liquidity \cite{shah-etal-2023-trillion,10.1145/3503161.3548380,ouyang-etal-2024-modal,RePEc:wsi:wsbook:6578,https://doi.org/10.1002/jae.3950070512,Zhou_Zhang_Peng_Zhang_Li_Xiong_Zhang_2021,pmlr-v162-zhou22g,Chen_2021_ICCV}. In Fig.~\ref{fig:event_demo}, we highlight several recent events where such announcements triggered pronounced S\&P~500 moves: the Fed's first post-pandemic 25-basis-point rate hike on March~16, 2022 sparked a relief rally (S\&P +2.24\%, Nasdaq +3.77\%); Chair Powell's ``far greater pain'' Jackson Hole speech on August~26, 2022 instead triggered a sharp risk-off move, with all three major U.S.\ indices falling by more than 3\%; another 25-basis-point hike on February~1, 2023, accompanied by Powell's first acknowledgement that disinflation had begun, lifted the S\&P by 1.05\% and the Nasdaq by 2.0\%; and a hotter-than-expected January~2024 CPI release led to a broad sell-off, with the S\&P losing 1.37\%, the Dow 1.35\%, and the 10-year Treasury yield jumping to about 4.31\%. Accurately estimating such event-driven market impacts is crucial not only for investors, who need to assess potential risks and returns, but also for policymakers, who must evaluate the consequences of their decisions.

With the rise of large language models (LLMs) and large time-series foundation models, recent forecasting research has advanced along three closely related directions. First, \emph{LLM-based time-series approaches} adapt pre-trained LLMs for prediction either by attaching task-specific predictive heads \cite{10.1145/3719207,NEURIPS2024_dcf88cbc,pan2024sipllm,yu-etal-2023-harnessing} or by carefully designing prompts \cite{cao2024tempo,xie2023wallstreetneophytezeroshot}. Second, a line of \emph{Transformer-based time-series models} leverages Transformer architectures for time-series forecasting, such as Informer \cite{Zhou_Zhang_Peng_Zhang_Li_Xiong_Zhang_2021}, FEDformer \cite{pmlr-v162-zhou22g}, and Autoformer \cite{Chen_2021_ICCV}, which are trained from scratch on forecasting tasks. More recently, large time-series foundation models, including MOMENT \cite{goswami2024moment}, Chronos \cite{ansari2024chronos}, and Kronos \cite{shi2025kronos}, extend this paradigm by pretraining on massive collections of time series to learn generic temporal representations, and have demonstrated strong performance across diverse datasets and tasks. However, these approaches are inherently unimodal, overlooking either textual information or time-series patterns. This has motivated a third direction: \emph{multimodal forecasting frameworks} that combine event text with time-series signals for prediction, such as GPT4MTS \cite{Jia_Wang_Zheng_Cao_Liu_2024}, Time-LLM \cite{jin2024timellm}, and TimeCMA \cite{Liu_Xu_Miao_Yang_Zhang_Long_Li_Zhao_2025}. Nevertheless, existing multimodal methods for event-driven forecasting still face three key limitations:

\paragraph{Limitation 1: Event-invariant fusion.}
Current multimodal forecasters typically fuse text and time-series representations via shared MLP layers \cite{10.1145/3711896.3736872} or attention blocks \cite{Jia_Wang_Zheng_Cao_Liu_2024,Liu_Xu_Miao_Yang_Zhang_Long_Li_Zhao_2025}, implicitly assuming a stable, event-independent balance between modalities that is reused across all events. However, event-driven markets are highly heterogeneous and non-stationary: some price movements are catalyzed by the sentiment and surprises in event scripts, while others are largely governed by pre-existing technical patterns. An event-invariant fusion scheme cannot adaptively reweight modalities across events, leading to suboptimal performance in heterogeneous market regimes.

\paragraph{Limitation 2: Lack of causal-aware fusion.}
Existing multimodal methods \cite{10.1145/3503161.3548380,ouyang-etal-2024-modal} do not incorporate explicit mechanisms that exploit the \emph{directional} influence of textual events on subsequent market reactions. They typically fuse modalities in a symmetric fashion and lack a principled, per-event criterion for deciding when event text is predictively useful beyond historical prices. Without such causal-aware fusion, these approaches may overuse noisy or uninformative text and fail to fully capture the fundamental drivers of market behavior, thereby limiting forecasting performance.

\paragraph{Limitation 3: Coarse-grained alignment.}
Existing alignment mechanisms are often instance-level, aligning a global event embedding with a global time-series embedding \cite{10.1145/3711896.3736872,10.1145/3503161.3548380,ouyang-etal-2024-modal}. In practice, only specific passages in financial releases are truly market-moving. For example, in FOMC statements, forward guidance on policy rates or the inflation outlook typically matters more than boilerplate language. Purely instance-level alignment can therefore be diluted by irrelevant or noisy content and fails to localize the fine-grained textual cues that drive market reactions.

\paragraph{Our contributions.}
To address these challenges, we propose \textbf{GS-Fuse}, a Granger-supervised and causal-aware multimodal framework for event-based financial forecasting. Our main contributions are:

\begin{itemize}
    \item[\ding{227}] \textbf{Granger-supervised, causal-aware gated fusion.} We design an event-adaptive gating module that leverages a Granger-style utility signal to decide when to trust event text. The gate is encouraged to open for events where incorporating the textual narrative yields incremental predictive value beyond a time-series-only baseline, and to close when the text is uninformative or harmful, leading to selective, utility-aware fusion.
    \item[\ding{227}]  \textbf{Multi-granularity cross-modal alignment.} We introduce a joint alignment scheme that operates both at the instance level (aligning global event and time-series embeddings) and at a finer granularity (linking salient textual cues to specific temporal patterns), enabling the model to localize which parts of the text account for observed market reactions.
    \item[\ding{227}] \textbf{Plug-and-play multimodal framework.} We provide a model-agnostic adapter that can wrap arbitrary LLMs and time-series foundation models. GS-Fuse requires only lightweight projection, fusion, and decoding modules, making it easy to instantiate on different backbones and market settings.
\end{itemize}

We evaluate GS-Fuse on a curated event dataset that integrates macroeconomic releases from CAMEF~\cite{10.1145/3711896.3736872} and salient financial-event news from FNSPID~\cite{10.1145/3637528.3671629}, and perform event-based forecasting for major equity indices and Treasury yields. Experiments show that GS-Fuse consistently outperforms strong unimodal and multimodal baselines, and ablations confirm the benefit of Granger-supervised, causal-aware fusion and multi-granularity alignment.

\section{Related Work}

\subsection{Event-Driven Financial Forecasting}
Event-driven financial forecasting \cite{BAO2025102616} studies how event signals such as macroeconomic releases, news, corporate announcements, and social media \cite{Gilbert2010-dt,liu2018leveragingfinancialnewsstock,zhou-etal-2021-trade,xu-cohen-2018-stock} affect asset prices and volatility \cite{RePEc:snb:snbwpa:2013-11,https://doi.org/10.1111/jofi.12196}. Existing work falls into three directions. 
(1) \textbf{Text-driven methods} model event semantics to predict market reactions, evolving from TF--IDF/topic models \cite{1196287,LI2014826,si-etal-2013-exploiting,NGUYEN20159603} to RNNs and pre-trained transformers \cite{10.1145/3155133.3155202,liu2018leveragingfinancialnewsstock,zhou-etal-2021-trade,shah-etal-2023-trillion}. 
(2) \textbf{Time-series-driven methods} focus on numerical modeling, from ARIMA/GARCH \cite{8212716,7046047,HYUPROH2007916} to deep RNN/CNN architectures and Transformer-based forecasters such as Informer and FEDformer \cite{liu2018leveragingfinancialnewsstock,8126078,Durairaj2022,Zhou_Zhang_Peng_Zhang_Li_Xiong_Zhang_2021,pmlr-v162-zhou22g}. Recently, time-series foundation models such as MOMENT, Timer, and TOKEN \cite{goswami2024moment,liu2024timer,anonymous2024totem} learn general, transferable temporal representations. 
(3) \textbf{Multi-modal methods} combine heterogeneous signals. Early work fused text and audio \cite{qin-yang-2019-say,10.1145/3366423.3380128}, and later studies integrated text and time series via SVM/GRU-style architectures \cite{10.1145/3394171.3413752,sawhney-etal-2020-deep}. These approaches often rely on relatively shallow temporal/semantic modeling, motivating recent transformer-based designs \cite{lee2024moat,Jia_Wang_Zheng_Cao_Liu_2024}. Our work follows this direction but leverages pre-trained encoders and a causally supervised fusion gate to better capture event impacts on market trends.

\vspace{-1em}
\subsection{Multi-Modal Alignment and Fusion}
In multimodal learning, \emph{alignment} maps modalities into a shared space, while \emph{fusion} integrates them for prediction \cite{li2025multimodalalignmentfusionsurvey}. Vision--language models move from instance-level contrastive alignment in CLIP \cite{pmlr-v139-radford21a} to token/patch-level matching in FILIP \cite{yao2022filip,Zhou_2022_CVPR}. By contrast, text--time-series models \cite{Liu_Xu_Miao_Yang_Zhang_Long_Li_Zhao_2025,10.1145/3711896.3736872,lee2024moat} largely operate at instance level and miss links between salient phrases and local temporal patterns. Our model jointly performs instance-level and token/step-level alignment to better ground event semantics in market trajectories. Gated fusion \cite{DBLP:conf/iclr/OvalleSMG17} weights modality-specific representations, and later work proposes more structured gates \cite{Cao_2023_ICCV,ganescu-etal-2025-looking,10.1145/3581783.3611805}. In event-driven financial forecasting, however, event text is an \emph{exogenous} driver of post-event prices, so gates trained only from end-task loss may track spurious correlations or the dominant modality. We instead introduce a Granger-supervised gate that ties fusion weights to the text modality's \emph{incremental} predictive gain.

\vspace{-1em}
\subsection{Deep Causality Learning}
Granger causality (GC) is a foundational framework for temporal causal analysis \cite{GRANGER1980329,10.1145/3762179}, stating that a source variable is causal to a target if adding its history improves the target's predictability. Neural GC methods extend this idea with nonlinear predictors and sparsity-based graph inference \cite{Khanna2020Economy}, and GC-inspired techniques have been applied to anomaly localization, spatiotemporal imputation, and learning under missing observations \cite{10.1145/3627673.3679642,Liu_Gao_Jiao_2025,Cheng_Li_Xiao_Li_Suo_He_Dai_2024}. Most prior work (i) operates in a single-modality setting and (ii) focuses on discovering causal graphs rather than exploiting causality for forecasting. In contrast, we propose a cross-modal Granger module that uses the causal modality's incremental predictive ability to supervise fusion and enhance event-driven forecasts.

\section{Preliminaries and Problem Formulation}
\label{sec:preliminaries}
We formalize event-driven financial forecasting, where a discrete macroeconomic release interacts with pre-event market dynamics to shape post-event price movements.

\subsection{Notation}
Let $\mathcal{D} = \{(E_i, X_i, Y_i)\}_{i=1}^N$ be a dataset of $N$ aligned multimodal instances:
\begin{itemize}
    \item \textbf{Event.} $E_i = (S_i, \tau_i, c_i)$, where $S_i = (w_{i,1}, \ldots, w_{i,m_i})$ is the event script, $\tau_i$ is its release time, and $c_i$ is the event category.
    \item \textbf{Pre-event market context.} A multivariate look-back window
    \begin{equation}
    X_i = (x_{\tau_i-L+1}, \ldots, x_{\tau_i}), \quad x_t \in \mathbb{R}^{d_x},
\end{equation}
where $L$ is the length of the pre-event window (number of historical time steps) and $x_t$ aggregates basic price/volume and technical features.

\item \textbf{Post-event target.} A future horizon segment
\begin{equation}
    Y_i = (y_{\tau_i+1}, \ldots, y_{\tau_i+H}), \quad y_t \in \mathbb{R}^{d_y},
\end{equation}
which represents the market response over the forecasting horizon $H$.
\end{itemize}

\begin{definition}[Event-driven causal response]
Let $E$ denote an event script, $X$ the pre-event market segment, and $Y$ the post-event segment. We treat the release as an exogenous information shock and posit the directed dependence
\begin{equation}
    (E, X) \rightarrow Y,
    \label{eq:causal-graph}
\end{equation}
so that $Y$ is jointly driven by salient information in $E$ and endogenous dynamics encoded in $X$.
\end{definition}

\begin{definition}[Event-driven forecasting task]
Given $(E_i, X_i)$ observed up to $\tau_i$, the goal is to learn a mapping
\begin{equation}
    f_\theta : (E_i, X_i) \mapsto \hat{Y}_i
\end{equation}
that approximates $Y_i$ over horizon $H$. Unlike standard time-series forecasting, which extrapolates from $X_i$ alone, event-driven forecasting must model the exogenous impact of $E_i$ together with the endogenous dynamics in $X_i$.
\end{definition}

\begin{figure*}[tp]
    \centering
    \includegraphics[width=\linewidth]{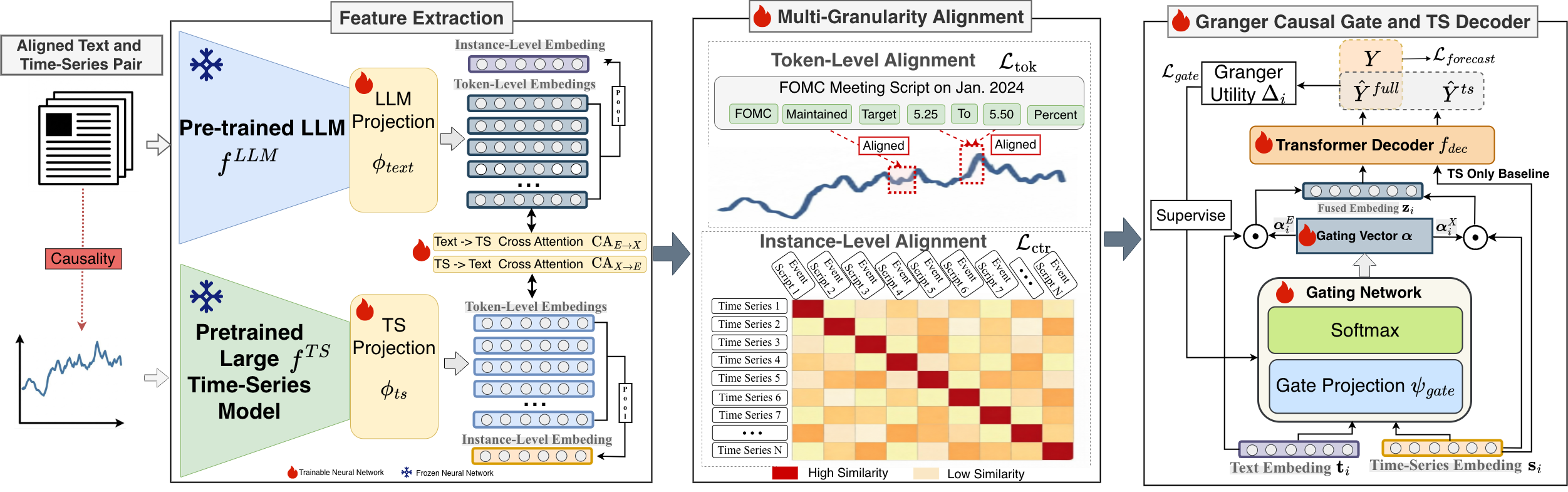}
    \caption{The Neural Architecture of GS-Fuse.}
    \label{fig:placeholder}
\end{figure*}

\vspace{-1em}

\section{Granger-Supervised Causal Gating}
\label{sec:gc_gate}
GS-Fuse supervises the fusion gate using a Granger-style criterion: the gate should open only when the event text provides incremental predictive value beyond a time-series-only baseline.

\subsection{Granger Causality Theory}
Granger causality formalizes the idea that a time series $X$ Granger-causes a target $Y$ if past values of $X$ improve forecasts of $Y$ beyond what can be achieved from the past of $Y$ alone \cite{GRANGER1980329}. Let
\begin{align}
    \hat{Y}^{\mathrm{res}}_{t+1} &= f_{\mathrm{res}}(Y_{1:t}), \\
    \hat{Y}^{\mathrm{unres}}_{t+1} &= f_{\mathrm{unres}}(Y_{1:t}, X_{1:t}),
\end{align}
denote a \emph{restricted} predictor and an \emph{unrestricted} predictor, respectively. Given a loss function $\ell(\cdot,\cdot)$, $X$ Granger-causes $Y$ if
\begin{equation}
    \mathbb{E}\big[\ell(Y_{t+1}, \hat{Y}^{\mathrm{unres}}_{t+1})\big] 
    < \mathbb{E}\big[\ell(Y_{t+1}, \hat{Y}^{\mathrm{res}}_{t+1})\big].
\end{equation}
In classical econometrics, this condition is evaluated via post-hoc statistical tests. In GS-Fuse, we reinterpret this principle as a \emph{Granger-supervision utility} that provides a local, instance-wise signal to regularize multimodal fusion.

\subsection{Granger-Supervised Gating}
Building on the classical framework, we extend the restricted-versus-unrestricted comparison to a multimodal setting. For each instance $i$, we compare a \emph{restricted} predictor that only observes the time-series modality $X_i$ and an \emph{unrestricted} predictor that additionally observes the event text $E_i$:
\begin{align}
    \hat{Y}^{\mathrm{ts}}_i   &= g^{\mathrm{ts}}_\theta(X_i), \\
    \hat{Y}^{\mathrm{full}}_i &= g^{\mathrm{full}}_\theta(X_i, E_i).
\end{align}
We define the \emph{Granger utility} of the text for instance $i$ as the loss differential:
\begin{equation}
    \Delta_i := \ell(Y_i, \hat{Y}^{\mathrm{ts}}_i) - \ell(Y_i, \hat{Y}^{\mathrm{full}}_i),
\end{equation}
where $\ell(\cdot,\cdot)$ is the forecasting loss (e.g., MSE) and $Y_i$ is the post-event target segment defined in \S~\ref{sec:preliminaries}.

\paragraph{Posterior Responsibility.}
To convert this utility into a supervision objective, we treat the predictors as competing experts and map their losses to soft responsibilities. Specifically, we pass the per-expert losses through a Gibbs-style transform
\[
p(Y_i \mid g_k) \propto \exp(-\ell_i^k/\tau_{\mathrm{gc}}), \quad k \in \{\mathrm{ts}, \mathrm{full}\},
\]
with $\ell_i^k = \ell(Y_i, \hat{Y}_i^k)$ and normalize them with a two-way softmax to obtain
\begin{equation}
    r_i = \frac{p(Y_i \mid g_{\mathrm{full}})}{p(Y_i \mid g_{\mathrm{full}}) + p(Y_i \mid g_{\mathrm{ts}})} = \sigma\left( \frac{\Delta_i}{\tau_{\mathrm{gc}}} \right),
\end{equation}
where $\sigma(\cdot)$ is the logistic sigmoid and $\tau_{\mathrm{gc}}$ is a temperature hyperparameter. Here, $r_i \in [0,1]$ serves as a soft ``text usefulness'' target: events where text yields significant predictive gains result in $r_i \approx 1$, while instances where text is redundant or noisy result in $r_i \approx 0$. This construction can be interpreted as a Bayesian posterior over the two experts under a Gibbs-style likelihood and a uniform prior.

\paragraph{Gate Supervision.}
We introduce a gating function $\alpha_i = \mathrm{Gate}(X_i, E_i) \in [0,1]$ that predicts this usefulness score directly from the inputs. The gate is trained to match $r_i$ via a distillation loss:
\begin{equation}
    \mathcal{L}_{\mathrm{gate}} = \frac{1}{N} \sum_{i=1}^N (\alpha_i - r_i)^2.
\end{equation}
Minimizing $\mathcal{L}_{\mathrm{gate}}$ encourages the gate to open (large $\alpha_i$) exactly on instances where the multimodal predictor exhibits positive Granger utility, allowing the fusion module to lean on the text branch. At inference time, $\alpha_i$ is computed solely from $(X_i, E_i)$; the Granger-based losses and $\Delta_i$ are used only during training.

\vspace{-0.5em}

\section{Model Architecture}
\label{sec:architecture}

Building on the problem formulation in \S~\ref{sec:preliminaries} and the Granger-supervised gating mechanism in \S~\ref{sec:gc_gate}, we now present the architecture of GS-Fuse. The model consists of four components: (i) a textual encoder for the event script $E_i$, (ii) a time-series encoder for the pre-event segment $X_i$, (iii) a Granger-gated fusion module with multi-granularity alignment, and (iv) a decoder that generates the future segment $Y_i$. All the hyperparameter values in this section are explained in Appendix B and Table \ref{tab:hyperparams}.

\subsection{Event and Time-Series Encoders}
\label{subsec:encoders}

\paragraph{Encoding event scripts via an LLM}
For each event $E_i=(S_i,\tau_i,c_i)$ with token sequence $S_i=(w_{i,1},\ldots,w_{i,m_i})$, we obtain token-wise text representations. A pretrained LLM $f^{\mathrm{LLM}}$ followed by a projection head $\phi_{\mathrm{text}}$ maps tokens into a shared space of dimension $F{=}1024$:
\begin{equation}
\mathbf{H}^{E}_i
=
\big(\mathbf{h}^{E}_{i,1},\ldots,\mathbf{h}^{E}_{i,m_i}\big)
=
\phi_{\mathrm{text}}\!\Big(f^{\mathrm{LLM}}(S_i)\Big)
\in \mathbb{R}^{m_i\times F}.
\end{equation}

\paragraph{Encoding time series via a TS foundation model.}
Given the pre-event segment $X_i=(x_{\tau_i-L+1},\ldots,x_{\tau_i})$, we extract step-wise representations. A pretrained time-series foundation encoder $f^{\mathrm{TS}}$ (e.g., MOMENT, Chronos) followed by a projection head $\phi_{\mathrm{ts}}$ maps the segment into the embedding space $F=1024$:
\begin{equation}
\mathbf{H}^{X}_i
=
\big(\mathbf{h}^{X}_{i,1},\ldots,\mathbf{h}^{X}_{i,L}\big)
=
\phi_{\mathrm{ts}}\!\Big(f^{\mathrm{TS}}(X_i)\Big)
\in \mathbb{R}^{L\times F}.
\end{equation}


We initialize $f^{\mathrm{LLM}}$ and $f^{\mathrm{TS}}$ from public foundation models and keep them frozen.
Concretely, for the textual modality $f^{\mathrm{LLM}}$ is instantiated from open-source LLM backbones (LLaMA-family and Phi-3, 3B) selected from the top ranked HuggingFace leaderboard \footnote{Huggingface LLM Leaderboard: https://huggingface.co/open-llm-leaderboard}, while for the time-series modality $f^{\mathrm{TS}}$ is instantiated from large time-series foundation models (MOMENT \cite{goswami2024moment} and Kronos \cite{shi2025kronos}), where MOMENT is among the first generic TS foundation models and Kronos is further tuned on stock-market data.
The projection heads $(\phi_{\mathrm{text}}, \phi_{\mathrm{ts}})$ and all downstream fusion/decoder modules in GS-Fuse are trained, while the backbone LLM and TS foundation models remain frozen across all experiments.

\vspace{-0.5em}

\subsection{Multi-Granularity Cross-Modal Alignment}
\label{sec:multi_align}
To reduce the modality gap and stabilize the joint representation space, we enforce pre-fusion alignment at both instance and token granularities.

\subsubsection{Instance-Level Global Alignment}
Given token/step representations $\mathbf{H}^{E}_i\in\mathbb{R}^{m_i\times F}$ and $\mathbf{H}^{X}_i\in\mathbb{R}^{L\times F}$, we first interleave information with bi-directional cross-attention:
\begin{align}
\tilde{\mathbf{H}}^{E}_i &= \mathrm{CA}_{X\rightarrow E}\!\left(\mathbf{Q}=\mathbf{H}^{E}_i,\ \mathbf{K}=\mathbf{H}^{X}_i,\ \mathbf{V}=\mathbf{H}^{X}_i\right),\\
\tilde{\mathbf{H}}^{X}_i &= \mathrm{CA}_{E\rightarrow X}\!\left(\mathbf{Q}=\mathbf{H}^{X}_i,\ \mathbf{K}=\mathbf{H}^{E}_i,\ \mathbf{V}=\mathbf{H}^{E}_i\right),
\end{align}
where $\mathrm{CA}(\cdot)$ is standard multi-head cross-attention. We then pool to obtain instance-level embeddings
\[
\mathbf{t}_i=\mathrm{Pool}(\tilde{\mathbf{H}}^{E}_i),\qquad
\mathbf{s}_i=\mathrm{Pool}(\tilde{\mathbf{H}}^{X}_i)\in\mathbb{R}^{F},
\]
and apply an InfoNCE loss over in-batch negatives:
\begin{equation}
    \mathcal{L}_{\mathrm{ctr}} = -\frac{1}{N}\sum_{i=1}^{N}
    \log \frac{\exp(\langle \hat{\mathbf{s}}_i, \hat{\mathbf{t}}_i \rangle / \tau_{\mathrm{ctr}})}
    {\sum_{k=0}^{N_{\mathrm{neg}}} \exp(\langle \hat{\mathbf{s}}_i, \hat{\mathbf{t}}_i^{(k)} \rangle / \tau_{\mathrm{ctr}})},
\end{equation}
where $\hat{\mathbf{s}}_i = \mathbf{s}_i / \|\mathbf{s}_i\|_2$ and 
$\hat{\mathbf{t}}_i^{(k)} = \mathbf{t}_i^{(k)} / \|\mathbf{t}_i^{(k)}\|_2$ denote $\ell_2$-normalized embeddings, $\mathbf{t}_i^{(0)}=\mathbf{t}_i$ is the positive pair, $\{\mathbf{t}_i^{(k)}\}_{k=1}^{N_{\mathrm{neg}}}$ are in-batch negatives, and $\tau_{\mathrm{ctr}}$ is the contrastive temperature.

\subsubsection{Token/Step Fine-Grained Alignment}
To capture local correspondences between salient phrases and specific temporal patterns, we further align individual text tokens with time-series steps.

\paragraph{Alignment space and soft positives.}
Let $\mathbf{H}_i^{E}\in\mathbb{R}^{m_i\times F}$ and $\mathbf{H}_i^{X}\in\mathbb{R}^{L\times F}$ be encoder hidden states. We map them into a shared alignment space:
\[
\mathbf{Z}_i^{E} = \mathrm{Norm}(\varphi^{E}(\mathbf{H}_i^{E})),\qquad
\mathbf{Z}_i^{X} = \mathrm{Norm}(\varphi^{X}(\mathbf{H}_i^{X}))\in\mathbb{R}^{L\times F},
\]
where $\varphi^{E},\varphi^{X}$ are learnable projections and $\mathrm{Norm}$ denotes row-wise $\ell_2$-normalization, so that $\mathbf{Z}_{i,j}^{E}$ and $\mathbf{Z}_{i,\ell}^{X}$ are unit-norm token and step embeddings, respectively. We form a token–step similarity matrix
\begin{equation}
    \mathbf{S}_i = \mathbf{Z}_i^{E}(\mathbf{Z}_i^{X})^\top / \tau_{\mathrm{al}}\in\mathbb{R}^{m_i\times L},
\end{equation}
with alignment temperature $\tau_{\mathrm{al}}$. An attention distribution from each token to time steps is
\begin{equation}
    \mathbf{P}_i^{E \rightarrow X} = \mathrm{softmax}_{\ell}(\mathbf{S}_i)\in\mathbb{R}^{m_i\times L},
\end{equation}
where $\mathrm{softmax}_{\ell}$ is applied row-wise over the time-step dimension $\ell$. The corresponding \emph{soft-positive} time-series vector for token $j$ is
\begin{equation}
    \mathbf{z}_{i,j}^{X,+} = \sum_{\ell=1}^{L} \mathbf{P}_i^{E \rightarrow X}(j, \ell)\,\mathbf{Z}_{i,\ell}^{X}.
\end{equation}

\paragraph{Salient anchors and contrastive loss.}
We score token salience with a linear scorer:
\begin{equation}
    a_{i,j} = \mathrm{softmax}_{j}\!\left(\mathbf{w}^\top \mathbf{h}^{E}_{i,j}\right),
\end{equation}
where $\mathbf{h}^{E}_{i,j}$ is the original text hidden state, $\mathbf{w}$ is learnable, and $\mathrm{softmax}_{j}$ is taken over tokens $j=1,\ldots,m_i$. The Top-$K_{\mathrm{top}}$ tokens under $a_{i,j}$ form an anchor set $\mathcal{A}_i$. For each anchor token $j\in\mathcal{A}_i$, we encourage its text embedding $\mathbf{Z}_{i,j}^{E}$ to match $\mathbf{z}_{i,j}^{X,+}$ and repel time-series steps from other instances via a salience-weighted InfoNCE loss with cross-sample negatives:
{\scriptsize
\begin{equation}
    \mathcal{L}_{\mathrm{tok}} = \frac{1}{N}\sum_{i=1}^{N}\sum_{j \in \mathcal{A}_i} a_{i,j}
    \left[
    -\log
    \frac{\exp(\langle \mathbf{Z}_{i,j}^{E}, \mathbf{z}_{i,j}^{X,+} \rangle / \tau_{\mathrm{nce}})}
    {\exp(\langle \mathbf{Z}_{i,j}^{E}, \mathbf{z}_{i,j}^{X,+} \rangle / \tau_{\mathrm{nce}}) + \sum_{i' \neq i,\, \ell} \exp(\langle \mathbf{Z}_{i,j}^{E}, \mathbf{Z}_{i',\ell}^{X} \rangle / \tau_{\mathrm{nce}})}
    \right],
\end{equation}
}
where $\tau_{\mathrm{nce}}$ is the NCE temperature and the sum over $(i',\ell)$ ranges over all time-series steps $\ell$ in other instances $i' \neq i$. The overall alignment loss is
\begin{equation}
    \mathcal{L}_{\mathrm{align}} = \mathcal{L}_{\mathrm{ctr}} + \mathcal{L}_{\mathrm{tok}},
\end{equation}
with $\mathcal{L}_{\mathrm{ctr}}$ aligning instance-level representations and $\mathcal{L}_{\mathrm{tok}}$ enforcing fine-grained token–step correspondences.

\begin{table*}[t]
\caption{Statistics of Dataset}
\label{tab:data_summary}
\begin{tabular}{@{}cc@{}}
\begin{minipage}{\columnwidth}
\centering
\caption*{A. Statistics of Event Script (Textual Data)}
\resizebox{\columnwidth}{!}{
\begin{tabular}{@{}lrrrr@{}}
\toprule
\textbf{Event Type}            & \textbf{\# Event Days} & \textbf{\# Articles} & \textbf{Avg. Length (words)} & \textbf{Coverage} \\ \midrule
FOMC decisions                 & 1482                   & 16,302               & 444                           & 2002--2024        \\
Employment situation           & 1494                   & 16,434               & 682                           & 2002--2024        \\
Unemployment insurance         & 2505                   & 27,729               & 647                           & 2002--2024        \\
CPI releases                   & 1905                   & 20,955               & 1,058                         & 2002--2024        \\
PPI releases                   & 2007                   & 22,357               & 792                           & 2002--2024        \\
GDP announcements              & 2874                   & 31,616               & 778                           & 2002--2024        \\ \bottomrule
\end{tabular}
}
\end{minipage}
&
\begin{minipage}{\columnwidth}
\centering
\caption*{B. Statistics of Financial Assets (Time-Series Data)}
\resizebox{\columnwidth}{!}{
\begin{tabular}{@{}l l r l l@{}}
\toprule
\textbf{Time Series Data}       & \textbf{Type}        & \textbf{\# Data Points} & \textbf{Range} & \textbf{Frequency}      \\ \midrule
S\&P500 index          & Equity index         & 331,256                 & 2008--2024     & 5-min                   \\
Dow Jones                & Equity index         & 187,241                 & 2012--2022     & 5-min                   \\
NASDAQ                & Equity index         & 332,615                 & 2009--2024     & 5-min                   \\
US 1M Treasury (USGG1M)         & Treasury yield       & 689,304                 & 2013--2024     & 5-min       \\
US 5Y Treasury (USGG5YR)        & Treasury yield       & 711,140                 & 2012--2024     & 5-min       \\ \bottomrule
\end{tabular}
}
\end{minipage}
\\
\end{tabular}
\end{table*}

\vspace{-0.5em}

\subsection{Granger-Gated Fusion}
\label{subsec:granger_gate}
We implement the Granger-supervised fusion principle from \S~\ref{sec:gc_gate}: the gate should open only when the event text provides incremental predictive utility beyond a TS-only baseline. Concretely, we produce a restricted (TS-only) forecast and an unrestricted (text-augmented) forecast using the \emph{same} decoder, and convert their loss gap into a stop-gradient supervision signal for the fusion gate.

\paragraph{Gating and fusion.}
Given instance embeddings $\mathbf{t}_i$ (event) and $\mathbf{s}_i$ (time series) from \S\ref{sec:multi_align}, the gating network takes their concatenation
\begin{align}
\mathbf{v}_i &= [\mathbf{t}_i;\mathbf{s}_i]\in\mathbb{R}^{2F}, \\
(\mathbf{a}^{E}_i,\mathbf{a}^{X}_i) &= \psi_{\mathrm{gate}}(\mathbf{v}_i), 
\qquad \mathbf{a}^{E}_i,\mathbf{a}^{X}_i\in\mathbb{R}^{F},
\end{align}
where $\psi_{\mathrm{gate}}$ is a small MLP. It produces feature-wise modality weights via a two-way softmax:
\begin{equation}
\alpha^{E}_{i,f}=\frac{\exp(a^{E}_{i,f}/\tau_{\mathrm{gate}})}
{\exp(a^{E}_{i,f}/\tau_{\mathrm{gate}})+\exp(a^{X}_{i,f}/\tau_{\mathrm{gate}})},
\qquad
\alpha^{X}_{i,f}=1-\alpha^{E}_{i,f},
\end{equation}
for $f=1,\ldots,F$, with temperature $\tau_{\mathrm{gate}}$. Let $\boldsymbol{\alpha}^{E}_i, \boldsymbol{\alpha}^{X}_i\in(0,1)^F$ and $\boldsymbol{\alpha}^{E}_i+\boldsymbol{\alpha}^{X}_i=\mathbf{1}$. The fused context is
\begin{equation}
\mathbf{z}_i=\boldsymbol{\alpha}^{E}_i\odot\mathbf{t}_i+\boldsymbol{\alpha}^{X}_i\odot\mathbf{s}_i\in\mathbb{R}^{F},
\end{equation}
where $\odot$ denotes element-wise multiplication. For gate supervision we use a scalar openness summary
\begin{equation}
\alpha_i \;=\; \frac{1}{F}\sum_{f=1}^{F}\alpha^{E}_{i,f}\;\in(0,1),
\end{equation}
i.e., the average responsibility assigned to the event modality.

\paragraph{Granger utility (restricted vs. unrestricted).}
To realize $g^{\mathrm{ts}}_\theta$ and $g^{\mathrm{full}}_\theta$ in \S\ref{sec:gc_gate} with a fair comparison, we reuse the same decoder $f_{\mathrm{dec}}$ (cf. \S\ref{subsec:decoder}) and compute
\begin{equation}
\hat{Y}^{\mathrm{full}}_i = f_{\mathrm{dec}}(\mathbf{z}_i), 
\qquad
\hat{Y}^{\mathrm{ts}}_i   = f_{\mathrm{dec}}(\mathbf{s}_i).
\end{equation}
We define per-instance losses (MSE over horizons and target dimensions)
\begin{equation}
\ell^{\mathrm{full}}_i=\mathrm{MSE}(\hat{Y}^{\mathrm{full}}_i, Y_i),\qquad
\ell^{\mathrm{ts}}_i=\mathrm{MSE}(\hat{Y}^{\mathrm{ts}}_i, Y_i),
\end{equation}
and obtain the Granger-supervision incremental utility
\begin{equation}
\Delta_i=\ell^{\mathrm{ts}}_i-\ell^{\mathrm{full}}_i,
\end{equation}
where $\Delta_i>0$ indicates that incorporating text reduces the forecasting loss beyond the TS-only baseline.

\paragraph{Responsibility target and gate supervision.}
We map the utility gap $\Delta_i$ to a soft responsibility target via a sigmoid:
\begin{equation}
    r_i = \sigma\!\big(\mathrm{clip}(\Delta_i / \tau_{\mathrm{gc}}, -c, c)\big),
\end{equation}
where $\sigma(\cdot)$ is the logistic sigmoid, $\tau_{\mathrm{gc}} > 0$ is a temperature hyperparameter, and $c > 0$ is a clipping threshold to avoid saturated gradients. Here, $r_i \in [0,1]$ serves as a soft ``text usefulness'' target: events where text yields significant predictive gains result in $r_i \approx 1$, while instances where text is redundant or noisy result in $r_i \approx 0$. We then supervise the gate by
\begin{equation}
    \mathcal{L}_{\mathrm{gate}} = \frac{1}{N}\sum_{i=1}^{N} (\alpha_i - r_i)^2,
\end{equation}
encouraging the gate to open when text provides reliable utility and to close otherwise. Finally, the training objective combines the forecasting loss with gate supervision:
\begin{equation}
    \mathcal{L} = \frac{1}{N}\sum_{i=1}^{N} \ell^{\mathrm{full}}_i \;+\; \lambda_{\mathrm{gate}}\,\mathcal{L}_{\mathrm{gate}},
\end{equation}
where $N$ is the batch size and $\lambda_{\mathrm{gate}}>0$ controls the strength of gate supervision.

\subsection{Time-Series Decoder}
\label{subsec:decoder}
Given the fused representation $\mathbf{z}_i \in \mathbb{R}^{F}$, we decode the future segment
$Y_i = (y_{\tau_i+1}, \ldots, y_{\tau_i+H})$
using a lightweight Transformer decoder with $L_{\mathrm{dec}}{=}3$ blocks and $n_{\mathrm{head}}{=}16$ attention heads.

\paragraph{Decoder input.}
We first expand $\mathbf{z}_i$ into a length-$H$ sequence by projecting it to the decoder dimension and adding horizon positional encodings:
\begin{equation}
    \mathbf{D}_i[h] = W_{\mathrm{in}}\mathbf{z}_i + \mathrm{PE}(h),
    \quad h = 1,\ldots,H,
\end{equation}
yielding $\mathbf{D}_i \in \mathbb{R}^{H \times d_{\mathrm{dec}}}$.

\paragraph{Transformer decoding.}
A stack of $L_{\mathrm{dec}}$ standard Transformer decoder blocks is applied to $\mathbf{D}_i$:
\begin{equation}
    \mathbf{H}_i = f_{\mathrm{dec}}(\mathbf{D}_i) \in \mathbb{R}^{H \times d_{\mathrm{dec}}},
\end{equation}
where $f_{\mathrm{dec}}$ denotes the Transformer decoder (composition of $L_{\mathrm{dec}}$ blocks).

\paragraph{Regression head.}
A small feed-forward head maps each horizon state to the target dimension $d_y$. For each step $h$,
\begin{equation}
    \hat{y}_{\tau_i+h} = W_{\mathrm{out}}\,\mathrm{MLP}_{\mathrm{reg}}(\mathbf{H}_i[h]) + \mathbf{b}_{\mathrm{out}},
\end{equation}
where $\mathrm{MLP}_{\mathrm{reg}}$ is a $K_{\mathrm{reg}}$-layer MLP with GELU activations. Collecting over $h$ yields the forecast
\begin{equation}
    \hat{Y}_i = (\hat{y}_{\tau_i+1}, \ldots, \hat{y}_{\tau_i+H}) \in \mathbb{R}^{H \times d_y}.
\end{equation}

\begingroup
\scriptsize
\renewcommand{\arraystretch}{0.86}
\setlength{\defaultaddspace}{0.2em}
\setlength{\aboverulesep}{0.4ex}
\setlength{\belowrulesep}{0.4ex}

\begin{table*}[tp]
\centering
\caption{Event-Driven Forecasting Performance. We report \textbf{MSE ($\times 10^{-4}$)}, \textbf{MAE ($\times 10^{-3}$)}, \textbf{DHR ($\times 10^{-2}$)}, and \textbf{Sharpe} on real-world financial datasets (S\&P500, NASDAQ, INDU, USGG1M, USGG5YR) across varying prediction horizons. \colorbox{gcfusecolor}{GS-Fuse} (Ours) and applied Time-Series Foundation Models \colorbox{foundationcolor}{Moment/Kronos} are highlighted. Best results are in \textbf{bold}.}
\label{tab:main_results}

\begin{adjustbox}{max width=\textwidth,center}
\begin{tabular}{@{}l|l|rrrr|rrrr|rrrr|rrrr|rrrr@{}}
\toprule
\multirow{2}{*}{\textbf{Model}} &
\multirow{2}{*}{\textbf{Horizon}} &
\multicolumn{4}{c|}{\textbf{S\&P500}} &
\multicolumn{4}{c|}{\textbf{NASDAQ}} &
\multicolumn{4}{c|}{\textbf{INDU}} &
\multicolumn{4}{c|}{\textbf{USGG1M}} &
\multicolumn{4}{c}{\textbf{USGG5YR}} \\
\cmidrule(lr){3-6}\cmidrule(lr){7-10}\cmidrule(lr){11-14}\cmidrule(lr){15-18}\cmidrule(lr){19-22}
& &
\textbf{MSE} & \textbf{MAE} & \textbf{DHR} & \textbf{Sharpe} &
\textbf{MSE} & \textbf{MAE} & \textbf{DHR} & \textbf{Sharpe} &
\textbf{MSE} & \textbf{MAE} & \textbf{DHR} & \textbf{Sharpe} &
\textbf{MSE} & \textbf{MAE} & \textbf{DHR} & \textbf{Sharpe} &
\textbf{MSE} & \textbf{MAE} & \textbf{DHR} & \textbf{Sharpe} \\
\midrule

& L=35
& 52.95 & 52.98 & 52.58 & 0.61
& 107.05 & 77.94 & 49.16 & 0.08
& 363.6 & 124.6 & 46.19 & -0.91
& 15.85 & 16.99 & 40.00 & -0.92
& 10.92 & 19.80 & 46.47 & -0.73 \\
& L=70
& 62.35 & 57.65 & -0.36 & 1.26
& 128.10 & 85.28 & 48.08 & 0.18
& 427.4 & 133.8 & 45.82 & -0.48
& 18.12 & 31.97 & 37.66 & -1.99
& 23.79 & 31.32 & 47.53 & -0.25 \\
\multirow{-3}{*}{\textbf{ARIMA}} & L=140
& 101.25 & 60.12 & 50.36 & -0.19
& 200.89 & 106.43 & 52.15 & 0.23
& 684.1 & 166.7 & 44.84 & -0.65
& 22.48 & 64.01 & 39.02 & -0.99
& 61.98 & 52.14 & 49.47 & -0.11 \\
\midrule

& L=35
& 81.36 & 68.78 & 45.27 & -1.18
& 110.68 & 79.69 & 53.46 & 0.71
& 357.2 & 127.6 & \textbf{55.95} & \textbf{2.61}
& 21.00 & 20.21 & \textbf{52.98} & \textbf{2.38}
& 13.57 & 25.01 & 46.47 & -0.98 \\
& L=70
& 100.29 & 77.45 & \textbf{54.01} & 1.13
& 186.35 & 105.96 & 54.84 & 0.90
& 539.6 & 162.8 & 55.37 & 0.59
& 26.97 & 42.28 & \textbf{70.85} & \textbf{4.35}
& 39.51 & 45.45 & 48.23 & -0.50 \\
\multirow{-3}{*}{\textbf{Dlinear}} & L=140
& 223.97 & 118.85 & 44.48 & -0.51
& 250.49 & 123.09 & 53.85 & 0.44
& 566.0 & 235.7 & 43.41 & 0.12
& 28.31 & 76.03 & 60.34 & 2.41
& 102.17 & 76.52 & 53.72 & 0.69 \\
\midrule

& L=35
& 55.95 & 55.18 & 50.86 & 1.08
& 110.20 & 79.29 & 53.61 & 0.12
& 333.4 & 125.0 & 50.24 & 1.35
& 17.50 & 21.30 & 40.85 & -1.01
& 66.92 & 66.80 & 47.00 & -1.28 \\
& L=70
& 73.57 & 63.40 & 50.43 & 1.13
& 145.67 & 92.23 & 54.69 & 1.42
& 417.1 & 135.8 & 49.64 & 0.49
& 24.19 & 42.44 & 59.57 & 0.83
& 28.50 & 35.54 & 51.41 & 0.37 \\
\multirow{-3}{*}{\textbf{Autoformer}} & L=140
& 115.67 & 78.18 & 51.94 & 0.14
& 255.20 & 121.84 & 51.08 & 0.49
& 759.4 & 180.3 & 46.28 & 0.21
& 26.90 & 85.15 & 62.05 & 1.46
& 72.78 & 59.80 & 48.05 & -0.23 \\
\midrule

& L=35
& 52.98 & 53.12 & 49.00 & -0.56
& 106.43 & 77.64 & 54.22 & 0.75
& 362.7 & 124.9 & 52.38 & 0.16
& 15.81 & 16.80 & 50.21 & 1.36
& 10.97 & 20.05 & 50.18 & 0.68 \\
& L=70
& 63.53 & 58.17 & 51.43 & -0.15
& 128.09 & 84.83 & 55.15 & 1.00
& 427.2 & 133.7 & 49.40 & -0.62
& 18.10 & 31.63 & 62.13 & 2.08
& 23.75 & 31.42 & 50.18 & 1.01 \\
\multirow{-3}{*}{\textbf{iTransformer}} & L=140
& 103.30 & 73.82 & 49.64 & -0.06
& 207.48 & 108.06 & 50.46 & -0.19
& 695.6 & 167.8 & 49.64 & -0.01
& 22.83 & 60.02 & 59.91 & 1.09
& 62.49 & 52.53 & 53.01 & 0.53 \\
\midrule

& L=35
& 52.82 & 53.01 & \textbf{54.29} & 0.81
& 107.42 & 78.00 & 50.38 & 0.61
& 107.4 & 78.00 & 51.58 & 0.84
& 16.59 & 19.57 & 40.00 & -2.30
& 10.72 & 19.99 & 46.29 & -0.65 \\
& L=70
& 65.40 & 60.17 & 53.46 & 0.68
& 140.15 & 89.68 & 49.92 & 0.80
& 135.6 & 88.20 & 49.43 & 0.57
& 25.87 & 43.19 & 49.79 & 0.09
& 28.86 & 37.70 & 48.94 & 0.76 \\
\multirow{-3}{*}{\textbf{FEDformer}} & L=140
& 115.96 & 78.48 & 49.16 & 0.42
& 233.03 & 115.41 & 51.38 & 0.78
& 681.7 & 168.2 & 52.37 & \textbf{0.95}
& 23.50 & 82.98 & 65.19 & 2.16
& 73.27 & 59.58 & 50.71 & 0.20 \\
\midrule

& L=35
& 34.40 & 33.05 & 50.57 & 1.16
& 42.62 & 43.78 & 51.31 & \textbf{1.69}
& 108.8 & 55.20 & 51.67 & 0.88
& 9.36 & 19.28 & 49.14 & 1.02
& 20.36 & 27.23 & 51.06 & 1.04 \\
& L=70
& 47.92 & 48.30 & 51.72 & 1.24
& 83.96 & 64.22 & 48.08 & 0.80
& 175.6 & 161.6 & 53.46 & 1.61
& 14.54 & 34.49 & 56.81 & 1.15
& 37.10 & 39.49 & 52.47 & 0.76 \\
\multirow{-3}{*}{\textbf{TimesFM}} & L=140
& 95.16 & 69.45 & 53.08 & 0.75
& 177.22 & 98.01 & 51.08 & 0.74
& 346.8 & 313.7 & 47.00 & -0.23
& 21.27 & 49.34 & 68.87 & 1.87
& 77.14 & 60.82 & 46.00 & -0.92 \\
\midrule

& L=35
& 21.19 & 30.43 & 53.70 & 0.34
& 38.73 & 42.10 & 52.79 & 0.62
& 118.4 & 61.80 & 50.14 & -0.19
& 8.97 & 15.31 & 47.28 & 1.05
& 17.84 & 26.50 & 48.68 & 0.16 \\
& L=70
& 42.75 & 44.39 & 53.59 & 1.00
& 77.44 & 60.96 & 51.47 & 0.56
& 227.7 & 88.90 & 50.37 & -0.42
& 13.73 & 27.35 & 57.06 & 2.07
& 34.62 & 37.84 & 49.34 & -0.30 \\
\multirow{-3}{*}{\textbf{Chronos}} & L=140
& 77.74 & 62.22 & 53.91 & 0.46
& 151.34 & 88.60 & 52.63 & 0.31
& 451.7 & 127.9 & 52.43 & -0.29
& 21.55 & 44.30 & 69.21 & 1.56
& 66.00 & 54.73 & 50.15 & -0.61 \\
\midrule

\rowcolor{foundationcolor} & L=35
& 22.19 & 30.86 & 49.98 & -0.00
& 41.14 & 42.98 & 49.68 & -0.08
& 135.3 & 66.20 & 50.07 & -0.17
& 9.28 & 16.03 & 47.88 & 1.68
& 18.61 & 27.07 & 49.49 & 0.46 \\
\rowcolor{foundationcolor} & L=70
& 43.86 & 45.23 & 49.72 & 0.02
& 81.34 & 63.06 & 50.00 & 0.05
& 268.6 & 97.40 & 50.13 & -0.25
& 13.80 & 27.37 & 53.71 & 2.14
& 36.53 & 39.17 & 49.48 & 0.25 \\
\rowcolor{foundationcolor} \multirow{-3}{*}{\textbf{Moment}} & L=140
& 85.06 & 64.77 & 52.37 & 0.44
& 159.01 & 90.41 & 50.36 & 0.14
& 538.6 & 140.2 & 51.17 & -0.09
& 19.93 & 44.01 & 66.12 & 2.58
& 70.52 & 56.57 & 48.93 & -0.04 \\
\midrule

\rowcolor{foundationcolor} & L=35
& 10.17 & 24.08 & 49.55 & 0.12
& 51.49 & 67.55 & 50.59 & 0.55
& 176.1 & 97.00 & 49.54 & -0.19
& 36.94 & 22.19 & 28.55 & 0.49
& 25.05 & 29.31 & 49.15 & 0.22 \\
\rowcolor{foundationcolor} & L=70
& 15.80 & 36.88 & 48.43 & -0.07
& 80.97 & 66.89 & 50.37 & -0.11
& 321.2 & 124.6 & 50.04 & -0.47
& 39.65 & 27.07 & 41.31 & 0.96
& 31.37 & 37.73 & 49.21 & 0.27 \\
\rowcolor{foundationcolor} \multirow{-3}{*}{\textbf{Kronos}} & L=140
& 85.39 & 90.82 & 51.69 & 0.26
& 178.01 & 104.73 & 51.21 & 0.02
& 666.2 & 466.1 & 50.38 & 0.43
& 53.57 & 61.05 & 37.74 & -1.57
& 79.77 & 87.48 & 50.12 & 0.28 \\
\midrule

& L=35
& 64.41 & 60.31 & 46.50 & -1.42
& 75.83 & 66.45 & 48.18 & -0.86
& 526.9 & 158.2 & 50.52 & 1.82
& 81.65 & 65.46 & 28.69 & 1.32
& 163.45 & 85.12 & 48.24 & 0.50 \\
& L=70
& 72.16 & 59.09 & 51.14 & 0.69
& 62.80 & 55.61 & 49.51 & 0.55
& 141.1 & 94.40 & 49.48 & 0.26
& 107.04 & 61.94 & 41.88 & 0.47
& 112.89 & 73.99 & 50.37 & 0.27 \\
\multirow{-3}{*}{\textbf{GPT4MTS}} & L=140
& 87.86 & 72.62 & 51.23 & 0.08
& 47.81 & 50.96 & 51.77 & -0.17
& 185.9 & 102.6 & 49.68 & 0.81
& 126.11 & 68.82 & 52.67 & 1.95
& 100.71 & 73.03 & 49.23 & 0.18 \\
\midrule

& L=35
& 13.65 & 17.01 & 46.30 & -0.61
& 12.54 & 15.07 & 43.82 & 0.20
& 33.6 & 46.60 & 52.59 & 2.28
& 36.49 & 37.01 & 19.80 & -1.81
& 34.43 & 37.94 & 51.26 & 1.14 \\
& L=70
& 10.69 & 24.71 & 50.49 & \textbf{1.36}
& 13.47 & 27.16 & 44.53 & -0.07
& 34.5 & 42.70 & 51.85 & 0.92
& 39.80 & 36.33 & 43.56 & 4.04
& 32.47 & 43.67 & 48.24 & 0.26 \\
\multirow{-3}{*}{\textbf{TEST}} & L=140
& 9.93 & 22.32 & 45.57 & -0.47
& 10.07 & 22.93 & 42.20 & -0.17
& 39.2 & 42.10 & 48.89 & -0.15
& 39.97 & 26.60 & 56.44 & 3.06
& 26.19 & 37.22 & \textbf{56.28} & \textbf{2.95} \\
\midrule

& L=35
& 32.43 & 39.60 & 49.47 & 0.16
& 59.96 & 55.21 & 48.64 & 0.27
& 187.6 & 82.00 & 50.00 & -0.59
& 9.16 & 21.31 & 49.48 & 2.27
& 27.37 & 34.40 & 49.44 & 0.50 \\
& L=70
& 65.14 & 58.74 & 48.52 & -0.06
& 78.80 & 62.60 & 49.90 & 0.11
& 394.6 & 124.6 & 50.90 & -0.20
& 13.80 & 30.71 & 52.84 & 1.55
& 37.28 & 40.22 & 49.16 & 0.18 \\
\multirow{-3}{*}{\textbf{TimeCMA}} & L=140
& 130.52 & 84.43 & 48.76 & -0.00
& 155.42 & 89.60 & 50.60 & 0.25
& 536.4 & 141.8 & 50.68 & 0.03
& 19.71 & 46.95 & 67.90 & 2.35
& 108.16 & 75.68 & 48.84 & -0.17 \\
\midrule[0.7pt]

\rowcolor{gcfusecolor} & L=35
& \textbf{3.38} & \textbf{11.44} & 53.16 & \textbf{1.92}
& \textbf{3.08} & \textbf{9.90} & \textbf{55.79} & 1.08
& \textbf{8.2} & \textbf{18.30} & 54.55 & 1.13
& \textbf{2.85} & \textbf{4.73} & 33.94 & 1.17
& \textbf{6.79} & \textbf{16.08} & \textbf{55.18} & \textbf{3.29} \\
\rowcolor{gcfusecolor} & L=70
& \textbf{3.98} & \textbf{12.48} & 53.84 & 1.25
& \textbf{3.66} & \textbf{10.80} & \textbf{59.82} & \textbf{1.78}
& \textbf{10.1} & \textbf{20.00} & \textbf{59.23} & \textbf{1.84}
& \textbf{3.13} & \textbf{5.24} & 68.11 & 2.77
& \textbf{8.94} & \textbf{19.19} & \textbf{58.23} & \textbf{2.88} \\
\rowcolor{gcfusecolor} \multirow{-3}{*}{\textbf{\begin{tabular}[c]{@{}l@{}}GS-Fuse\\ (LLaMa + MOMENT)\end{tabular}}} & L=140
& \textbf{6.60} & \textbf{15.67} & \textbf{56.03} & \textbf{0.79}
& \textbf{6.86} & \textbf{15.49} & \textbf{58.78} & \textbf{0.87}
& \textbf{18.2} & \textbf{25.00} & \textbf{61.40} & 0.86
& \textbf{3.84} & \textbf{6.89} & \textbf{73.96} & \textbf{3.27}
& \textbf{12.45} & \textbf{23.17} & 54.19 & 1.44 \\
\bottomrule
\end{tabular}
\end{adjustbox}
\end{table*}

\endgroup

\subsection{Three-Stage Training Scheme and Objectives}
\label{sec:three_stage_training}
We train \textsc{GS-Fuse} in three stages for stable optimization.

\paragraph{Stage 1: TS-only pre-training.}
We first train the time-series branch in a sliding-window setting: given segments of length $L$, the model predicts the next $H$ steps. Text inputs and fusion modules are omitted; only TS-side projections and the decoder are updated. The forecasting loss is
\begin{equation}
    \mathcal{L}_{\mathrm{forecast}}
    = \frac{1}{N} \sum_{i=1}^{N} \mathrm{MSE}(\hat{Y}_i, Y_i)
    + \frac{1}{N} \sum_{i=1}^{N} \mathrm{MAE}(\hat{Y}_i, Y_i),
\end{equation}
where $\hat{Y}_i$ is the decoded multi-horizon prediction for instance $i$ and $Y_i$ is the ground-truth trajectory.

\paragraph{Stage 2: Text-only warm-up.}
We then warm up the textual branch on the event-aligned dataset, using the same $\mathcal{L}_{\mathrm{forecast}}$. The time-series branch and fusion modules are bypassed; we update only text-side projections and the decoder so that event representations become directly forecasting-relevant.

\paragraph{Stage 3: Multimodal fine-tuning.}
Finally, we enable the full \textsc{GS-Fuse} model and jointly train with both modalities. The LLM and TS foundation encoders are frozen, while the fusion stack (cross-modal alignment and Granger-gated fusion) and prediction heads are trained with the combined objective
\begin{equation}
\mathcal{L}_{\mathrm{total}}
    = \mathcal{L}_{\mathrm{forecast}}
    + \lambda_{\mathrm{align}} \mathcal{L}_{\mathrm{align}}
    + \lambda_{\mathrm{gate}} \mathcal{L}_{\mathrm{gate}},
\end{equation}
where $\mathcal{L}_{\mathrm{align}}$ and $\mathcal{L}_{\mathrm{gate}}$ are defined in \S~\ref{sec:multi_align} and \S~\ref{sec:gc_gate}, and $\lambda_{\mathrm{align}}, \lambda_{\mathrm{gate}}$ are scalar weights.

\begin{table}[tp]
\centering
\caption{Backbone interaction on long-horizon forecasting ($L{=}140$). Metrics: \textbf{MSE ($\times10^{-4}$)} and \textbf{MAE ($\times10^{-3}$)}. $\Delta$ denotes the relative change when swapping the LLaMA text backbone for Phi; \textcolor{green!60!black}{$\downarrow$} improves and \textcolor{red}{$\uparrow$} degrades.}
\label{tab:interaction_analysis}

\resizebox{0.8\columnwidth}{!}{
\begin{tabular}{l ccc ccc}
\toprule
\multirow{2.5}{*}{\textbf{Foundation}} & \multicolumn{3}{c}{\textbf{MSE ($10^{-4}$)}} & \multicolumn{3}{c}{\textbf{MAE ($10^{-3}$)}} \\
\cmidrule(lr){2-4} \cmidrule(lr){5-7}
 & \textbf{LLaMA} & \textbf{Phi} & \textbf{Gap ($\Delta$)} & \textbf{LLaMA} & \textbf{Phi} & \textbf{Gap ($\Delta$)} \\
\midrule
\multicolumn{7}{l}{\textit{\textbf{(a) Dataset: INDU}}} \\
\cmidrule(lr){1-7}
\textbf{Moment} & \textbf{18.2} & 20.3 & \textcolor{red}{$\uparrow$ 11.5\%} & \textbf{25.0} & 26.6 & \textcolor{red}{$\uparrow$ 6.4\%} \\
\textbf{Kronos} & 25.9 & \textbf{25.4} & \textcolor{green!60!black}{$\downarrow$ 1.9\%} & 29.0 & \textbf{28.2} & \textcolor{green!60!black}{$\downarrow$ 2.8\%} \\

\midrule
\multicolumn{7}{l}{\textit{\textbf{(b) Dataset: USGG1M}}} \\
\cmidrule(lr){1-7}
\textbf{Moment} & \textbf{3.84} & 4.06 & \textcolor{red}{$\uparrow$ 5.7\%} & \textbf{6.89} & 7.33 & \textcolor{red}{$\uparrow$ 6.4\%} \\
\textbf{Kronos} & 4.09 & \textbf{3.97} & \textcolor{green!60!black}{$\downarrow$ 2.9\%} & 8.09 & \textbf{7.38} & \textcolor{green!60!black}{$\downarrow$ 8.8\%} \\
\bottomrule
\end{tabular}
}
\end{table}

\vspace{-1em}

\section{Experiments}
We evaluate \textsc{GS-Fuse} on event-driven market forecasting and address four research questions:
\textbf{RQ1 (Accuracy).} How well does \textsc{GS-Fuse} forecast markets across datasets and horizons?
\textbf{RQ2 (Generality).} How robust is \textsc{GS-Fuse} with different LLM text and time-series foundation backbones?
\textbf{RQ3 (Ablation).} Which components contribute most to performance?
\textbf{RQ4 (Causal Gating).} Does the Granger gate increase reliance on text when it helps and down-weight it otherwise?
We also provide case studies on aligned tokens from token-wise alignment module in Appendix D.

\subsection{Experimental Setting}

\subsubsection{Dataset and Curation}
\label{sec:dataset}
\paragraph{Event Script Set.}
We start from the six categories of salient macroeconomic events curated in the \textsc{CAMEF} dataset \cite{10.1145/3711896.3736872}, monetary policy decisions, employment reports (employment situation and unemployment insurance), inflation releases (CPI and PPI), and GDP announcements.\footnote{\textbf{CAMEF Dataset}: https://github.com/lakebodhi/CAMEF} To enrich each category, we retrieve additional financial news from \textsc{FNSPID}\footnote{\textbf{FNSPID Dataset}: https://github.com/Zdong104/FNSPID\_Financial\_News\_Dataset} using Sentence-BERT embeddings: an article is retained if its cosine similarity to an event-type prototype (mean-pooled \textsc{CAMEF} articles of that type) exceeds 0.55, with edge cases manually checked.

\paragraph{Financial Time-Series Data.}
We directly use the high-frequency market data from \textsc{CAMEF}, which provides 5-minute trading series for three major U.S. equity indices (S\&P500, Dow Jones Industrial Average, NASDAQ) and two U.S. Treasury bonds (1-month and 5-year). These assets are well documented as sensitive to macroeconomic announcements.

The train, validation, and test sets are split in a 6 / 2 / 2 ratio over time: the early events and time series are used for training and validation, and the latest data are reserved for testing.

\subsubsection{Test Setting}
\label{sec:test_setting}
Following prior work on macro announcement effects \cite{https://doi.org/10.1111/jofi.12818,GERTLER2018336,https://doi.org/10.1111/joes.12550,RePEc:ijc:ijcjou:y:2016:q:4:a:6,TADLE2022106021,RePEc:fip:fednep:00004,https://doi.org/10.1111/jofi.12196,ROSA2011915}, we assume that event information is incorporated into prices within at most two trading days. We therefore evaluate three forecasting horizons: half a trading day, one full trading day, and two trading days, which at 5-minute frequency correspond to horizon lengths $H \in \{35, 70, 140\}$. As in \S~\ref{sec:preliminaries}, we set the look-back length equal to the horizon and use $L = H$ in all experiments.

\subsubsection{Implementation Overview}
For the single-modality time-series baselines (ARIMA, DLinear, Autoformer, FEDformer, iTransformer), we train each model on continuous historical data in a sliding-window fashion: given a past segment $X_{t-L+1:t}$, the model predicts the next $H$ steps $X_{t+1:t+H}$, with $L = H \in \{35, 70, 140\}$. At evaluation time, for an event $i$ released at timestamp $\tau_i$, we extract its pre-event window $X_i = (x_{\tau_i - L + 1}, \dots, x_{\tau_i})$ and forecast the subsequent segment $(x_{\tau_i + 1}, \dots, x_{\tau_i + H})$. For multimodal approaches (GPT4MTS, TEST, TimeCMA, and \textsc{GS-Fuse}), we use the same event-aligned windows but provide both the event script $E_i$ and pre-event context $X_i$ as input to predict the post-event trajectory over horizon $H$. We provide detailed implementations in Appendices A and B.

\begin{figure}[tp]
    \centering
    
    \includegraphics[width=\columnwidth]{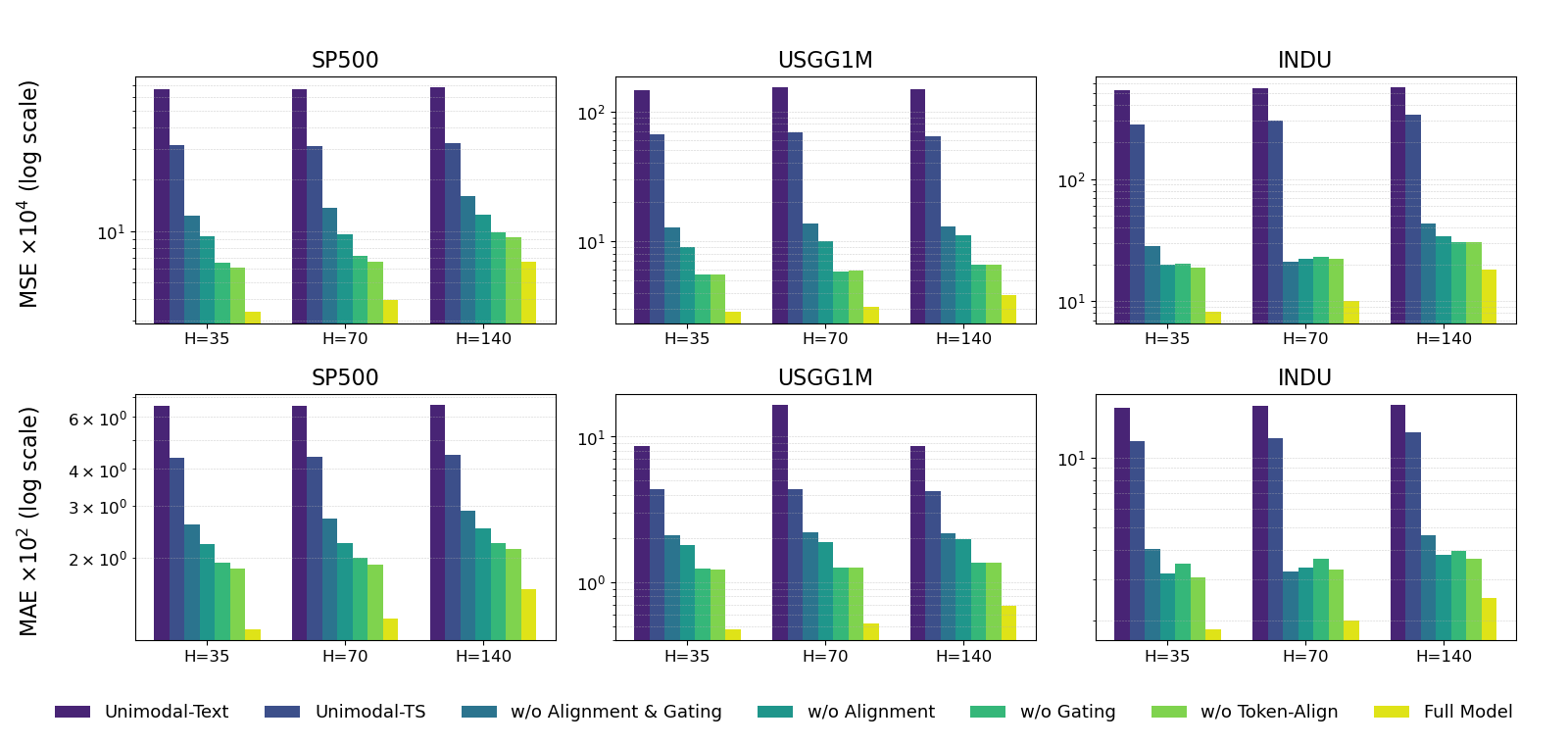}
    \caption{Ablation of GS-Fuse components. We compare text-only and time-series-only variants
(\textit{Unimodal-Text}, \textit{Unimodal-TS}) with multimodal variants that progressively remove
Granger-gated fusion and instance- and token-level alignment
(\textit{w/o Alignment \& Gating}, \textit{w/o Alignment}, \textit{w/o Gating},
\textit{w/o Token-Align}, \textit{Full Model}). Bars show MSE (top) and MAE (bottom); lower is better.}
\label{fig:ablation}

\end{figure}

\vspace{-0.5em}

\subsection{Event-based Market Trend Forecasting (RQ1)}



To answer \textbf{RQ1}, we compare GS-Fuse with twelve statistical, neural, and foundation-model baselines on five financial datasets (S\&P500, INDU, NASDAQ, USGG1M, USGG5YR) over three horizons ($L \in \{35, 70, 140\}$), reporting MSE and MAE in Table~\ref{tab:main_results}. Across all assets and horizons, GS-Fuse achieves the lowest errors, substantially improving over both time-series foundation models (e.g., on S\&P500 with $L=35$, GS-Fuse reduces MSE from 1.017 for Kronos to 0.338) and multimodal baselines such as GPT4MTS and TimeCMA. These gains are particularly pronounced when event text is noisy or boilerplate, where standard cross-attention fusion can overfit irrelevant tokens, while our Granger-supervised gate only incorporates text when it provides measurable incremental utility beyond the TS-only branch. GS-Fuse also remains stable as the horizon extends to two trading days and generalizes across equity indices and Treasury yields, with MAE exhibiting the same pattern of improvements as MSE. The additional DHR and Sharpe results further confirm that GS-Fuse improves not only point-wise forecasting accuracy, but also directional prediction and risk-adjusted trading utility.

\subsection{Ablation Study on Gating and Alignment (RQ3)}

Fig. \ref{fig:ablation} illustrates that both unimodal variants (\emph{Unimodal-Text} and \emph{Unimodal-TS}) perform markedly worse than all multimodal configurations, confirming that jointly modeling event scripts and pre-event time series is essential. Among multimodal variants, removing both Granger-gated fusion and alignment (\emph{w/o Alignment \& Gating}) yields the largest performance drop, while reintroducing either the gate (\emph{w/o Alignment}) or the multi-granularity alignment (\emph{w/o Gating}) recovers a substantial portion of the lost accuracy. Further enabling token-level alignment (\emph{w/o Token-Align}) brings consistent additional gains, and the full GS-Fuse model attains the lowest MSE and MAE across all settings, indicating that Granger-supervised gating and instance/token-level alignment are complementary and jointly responsible for the overall improvements.

\begin{table}[tp]
    \centering
    \caption{Internal comparison between the TS-only branch and the full GS-Fuse model. We report MSE ($\times 10^{-4}$) and MAE ($\times 10^{-3}$) and the relative improvement ($\Delta$). The "Avg. Gate" row indicates the model's reliance on the text branch.}
    \label{tab:ablation_branch}
    \resizebox{0.9\columnwidth}{!}{
    \begin{tabular}{l cc cc cc}
        \toprule
        \multirow{2}{*}{\textbf{Method}} & \multicolumn{2}{c}{\textbf{INDU}} & \multicolumn{2}{c}{\textbf{S\&P500}} & \multicolumn{2}{c}{\textbf{USGG1M}} \\
        \cmidrule(lr){2-3} \cmidrule(lr){4-5} \cmidrule(lr){6-7}
         & MSE & MAE & MSE & MAE & MSE & MAE \\
        \midrule
        TS-only Branch & 18.53 & 25.30 & 6.70 & 15.82 & 3.99 & 8.05 \\
        \textbf{Full Model} & \textbf{18.16} & \textbf{24.99} & \textbf{6.60} & \textbf{15.67} & \textbf{3.84} & \textbf{6.89} \\
        \emph{Improv. ($\Delta$ \%)} & \emph{+2.0\%} & \emph{+1.2\%} & \emph{+1.5\%} & \emph{+1.0\%} & \emph{+3.8\%} & \emph{+14.4\%} \\
        \midrule
        Avg. Text Gate & \multicolumn{2}{c}{52.89\%} & \multicolumn{2}{c}{51.55\%} & \multicolumn{2}{c}{60.68\%} \\
        \bottomrule
    \end{tabular}
    }
    \vspace{-1em}
\end{table}

\vspace{-0.5em}

\subsection{Backbone Generality: Performance Across LLM and Time-Series Encoders (RQ2)}

To address \textbf{RQ2} (backbone generality), we instantiate GS-Fuse with two compact LLMs, namely LLaMA-3B and Phi-3B, chosen from top-ranked models on the Hugging Face LLM Leaderboard\footnote{Hugging Face LLM Leaderboard: \url{https://huggingface.co/open-llm-leaderboard}} that remain trainable on a single RTX 4090 GPU, and with two leading time-series foundation models, MOMENT (broad multi-domain pretraining) and Kronos (finance-focused pretraining). Table~\ref{tab:interaction_analysis} reports long-horizon ($L=140$) results on INDU and USGG1M and reveals clear interaction effects: MOMENT pairs best with LLaMA (switching to Phi consistently worsens both MSE and MAE), whereas Kronos slightly prefers Phi (yielding small but consistent error reductions). All four combinations remain strong and outperform time-series-only baselines, indicating that GS-Fuse is backbone-agnostic but not backbone-independent: carefully matching LLM and time-series foundations can provide an additional $\sim$5--10\% improvement on top of the gains established in Table~\ref{tab:interaction_analysis}.

\vspace{-0.5em}

\subsection{Analysis on Granger-Supervised Gating (RQ4)}

As shown in Table~\ref{tab:ablation_branch}, GS-Fuse consistently improves over its internal TS-only branch across the three assets, including the stock indices INDU, S\&P500 and the treasury bond USGG1M, yielding $1.5$--$2.0\%$ MSE reductions on INDU and S\&P500 and a larger $3.8\%$ MSE and $14.4\%$ MAE reduction on USGG1M. The averaged text-gate values are roughly balanced on the equity indices ($\approx 51\%$), but noticeably higher on USGG1M ($60.68\%$), indicating that the model relies on macro texts more frequently when forecasting short-term Treasury yields. These results support the Granger-supervised design: GS-Fuse preserves the strong TS baseline while selectively leveraging textual information where it provides greater incremental predictive value. Appendix C provides additional gate analyses.

\vspace{-0.5em}

\section{Conclusion}
We introduced \emph{GS-Fuse}, a Granger-supervised causal fusion framework that models how macro event texts interact with market time series via gated fusion and multi-granularity alignment. Across five major assets and multiple horizons, GS-Fuse outperforms strong baselines, and gate analyses show it relies on text only when event narratives add predictive value beyond prices, pointing to a solid foundation for future extensions to richer markets and decision-centric applications.

\section*{Acknowledgements}

This work was supported in part by the National Natural Science Foundation of China (NSFC) under Grant Nos. 62402396 and 72471197. We would like to thank the anonymous reviewers for their valuable comments and suggestions that helped improve the quality of this paper.

\bibliographystyle{ACM-Reference-Format}
\bibliography{references}

\appendix

\section{Implementation Details for Baselines and GS-Fuse}

Single-modality baselines (ARIMA, DLinear, Autoformer, FEDformer, iTransformer, PatchTST) are adapted from Time-Series-Library \footnote{Time-Series-Library: \url{https://github.com/thuml/Time-Series-Library}} to the event-driven setting, using pre-event segments to predict 35-, 70-, and 140-step horizons. All these baselines are trained for 10 epochs with batch size 32.

As multimodal baselines, we use TEST \footnote{TEST: \url{https://github.com/SCXsunchenxi/TEST}}, GPT4MTS\footnote{GPT4MTS: \url{https://github.com/Flora-jia-jfr/GPT4MTS-Prompt-based-Large-Language-Model-for-Multimodal-Time-series-Forecasting}}, and TimeCMA \footnote{TimeCMA: \url{https://github.com/ChenxiLiu-HNU/TimeCMA}}. We adapt their official implementations to the same event-driven forecasting horizons (35, 70, and 140 steps) and keep their default hyperparameters.

For time-series foundation models, including MOMENT, Kronos, TimesFM, and Chronos, we use the official repositories\footnote{%
  MOMENT: \url{https://github.com/moment/moment}\\
  Kronos: \url{https://github.com/shiyu-coder/Kronos}\\
  TimesFM: \url{https://github.com/google-research/timesfm}\\
  Chronos: \url{https://github.com/amazon-science/chronos-forecasting}%
}
and retrain them on our event-driven setup for 10 epochs with batch size 32.

For \textsc{GS-Fuse}, we adopt the three-stage training paradigm described in \S\ref{sec:architecture}, training each stage for 2 epochs with batch size 32.

\section{Hyperparameter Setting of GS-Fuse}

All hyperparameters are fixed across datasets. The shared fusion dimension is set to \(F=1024\) to match the typical scale of LLM and time-series foundation model embeddings and to provide sufficient capacity for cross-modal interactions without making the model excessively large. The look-back length \(L\) and forecast horizon \(H\) follow standard settings in event-driven forecasting (roughly one pre-event day and up to two post-event days) so that our setup is comparable to prior work. For the backbone encoders, we select compact open-weight LLMs (LLaMA / Phi-3 family) and time-series foundation models (MOMENT / Kronos) as a pragmatic trade-off between representation quality and computational cost.

For multi-granularity alignment, we adopt conventional contrastive-learning temperatures and anchor sizes. The instance-level and token-level temperatures \(\tau_{\text{ctr}}, \tau_{\text{al}}, \tau_{\text{nce}}\) are chosen in the typical range used in contrastive and vision--language models (on the order of \(10^{-2}\)–\(10^{-1}\)), which yields non-degenerate softmax distributions while keeping gradients stable. The number of in-batch negatives \(N_{\text{neg}}\) and the top-\(K\) token anchors are set to \(K_{\text{top}} = 256\), which typically covers about 30\%–50\% of tokens in our event scripts. This provides a balance between capturing salient spans and avoiding excessive influence from noisy or redundant tokens.

For the Granger-supervised gate, we use simple, stable defaults. The gating temperature \(\tau_{\text{gate}} = 1.0\), gain \(\gamma = 1.0\), minimum scale \(\epsilon\), and clipping threshold \(c\) are set so that the responsibility signal \(r_i\) meaningfully reflects the error gap between the time-series-only and multimodal branches while keeping logits in a numerically safe range. The resulting effective Granger temperature \(\tau_{\text{gc}} = s_\Delta / \gamma\) can be viewed as an automatic rescaling of this gap; this mechanism is kept fixed rather than tuned per dataset.

The decoder configuration is intentionally lightweight. A 3-layer Transformer decoder with \(n_{\text{head}} = 16\) and hidden size \(d_{\text{dec}} = F\) (i.e., 64 dimensions per head) provides a good balance between expressiveness and runtime. The regression head is a 2-layer MLP with GELU activations, a standard choice in Transformer architectures that introduces mild nonlinearity without many extra parameters.

Finally, the loss weights are set to keep forecasting as the primary objective. The alignment and gate losses are down-weighted by \(\lambda_{\text{align}} = 0.2\) and \(\lambda_{\text{gate}} = 0.1\), respectively, so that they regularize the model toward consistent cross-modal representations and Granger-consistent gating but do not overwhelm the core MSE+MAE forecasting loss. These settings are intentionally simple, fixed across all experiments, and consistent with the qualitative role of each component.

\vspace{-1em}

\begin{table}[t]
\caption{Hyperparameters of \textsc{GS-Fuse} used in experiments.}
\vspace{-1em}
\label{tab:hyperparams}
\centering
\resizebox{\columnwidth}{!}{
\begin{tabular}{lllp{4.8cm}}
\toprule
\textbf{Module} & \textbf{Symbol} & \textbf{Value} & \textbf{Role} \\
\midrule
\multirow{5}{*}{Encoders}
  & $F$              & 1024                      & Shared text--time-series fusion dimension. \\
  & $L$              & task-dependent            & Look-back length of the pre-event time series. \\
  & $H$              & task-dependent            & Forecast horizon length. \\
  & $f_{\text{LLM}}$ & LLaMA / Phi-3 family      & Backbone text encoder for event scripts. \\
  & $f_{\text{TS}}$  & MOMENT / Kronos           & Time-series foundation encoder for market context. \\
\midrule
\multirow{5}{*}{Alignment}
  & $\tau_{\text{ctr}}$ & tuned                  & Temperature in instance-level InfoNCE loss $L_{\text{ctr}}$. \\
  & $N_{\text{neg}}$    & in-batch              & Number of negatives per anchor in $L_{\text{ctr}}$. \\
  & $\tau_{\text{al}}$  & 0.2                   & Temperature in token--step similarity matrix $S_i$. \\
  & $\tau_{\text{nce}}$ & 0.07                  & Temperature in token-level InfoNCE loss $L_{\text{tok}}$. \\
  & $K_{\text{top}}$    & 256                   & Number of top-salience text tokens forming anchor set $A_i$. \\
\midrule
\multirow{5}{*}{Granger gate}
  & $\tau_{\text{gate}}$ & 1.0                  & Softmax temperature for feature-wise modality gating. \\
  & $\epsilon$           & $10^{-6}$            & Minimum scale in $s_\Delta = \max\!\big(\frac{1}{N}\sum_i |\Delta_i|,\epsilon\big)$. \\
  & $\gamma$             & 1.0                  & Gain factor in responsibility target $r_i$. \\
  & $c$                  & 6.0                  & Clipping threshold for scaled utility before sigmoid. \\
  & $\tau_{\text{gc}}$   & $s_\Delta / \gamma$  & Effective Granger temperature induced by scaling. \\
\midrule
\multirow{5}{*}{Decoder}
  & $L_{\text{dec}}$   & 3                      & Number of Transformer decoder blocks. \\
  & $n_{\text{head}}$  & 16                     & Attention heads per decoder block. \\
  & $d_{\text{dec}}$   & $F$ (1024)             & Decoder hidden dimension. \\
  & $K_{\text{reg}}$   & 2                      & Depth of regression head MLP. \\
  & $\rho(\cdot)$      & GELU                   & Activation function in the regression head. \\
\midrule
\multirow{2}{*}{Loss weights}
  & $\lambda_{\text{align}}$ & 0.2             & Weight of alignment loss $L_{\text{align}}$ in the total objective. \\
  & $\lambda_{\text{gate}}$  & 0.1             & Weight of gate loss $L_{\text{gate}}$ in the total objective. \\
\bottomrule
\end{tabular}
}
\vspace{-2em}
\end{table}
\begin{figure*}[t]
    \centering
    \begin{subfigure}[t]{0.48\textwidth}
        \centering
        \includegraphics[width=\textwidth]{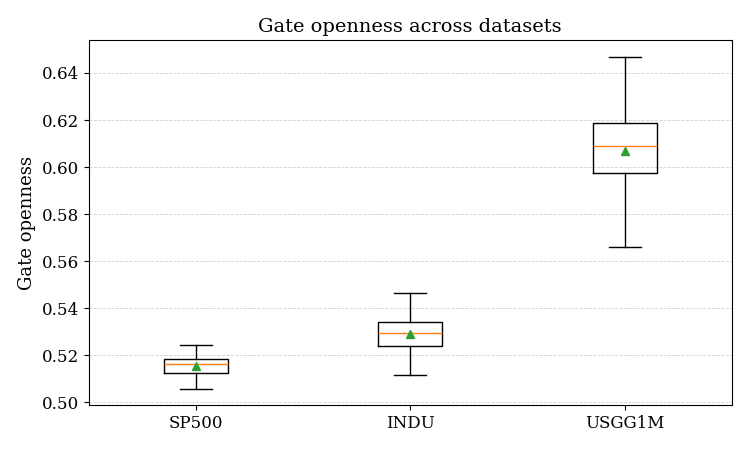}
        \caption{Gate openness across datasets.}
        \label{fig:gate_datasets}
    \end{subfigure}
    \hfill
    \begin{subfigure}[t]{0.48\textwidth}
        \centering
        \includegraphics[width=\textwidth]{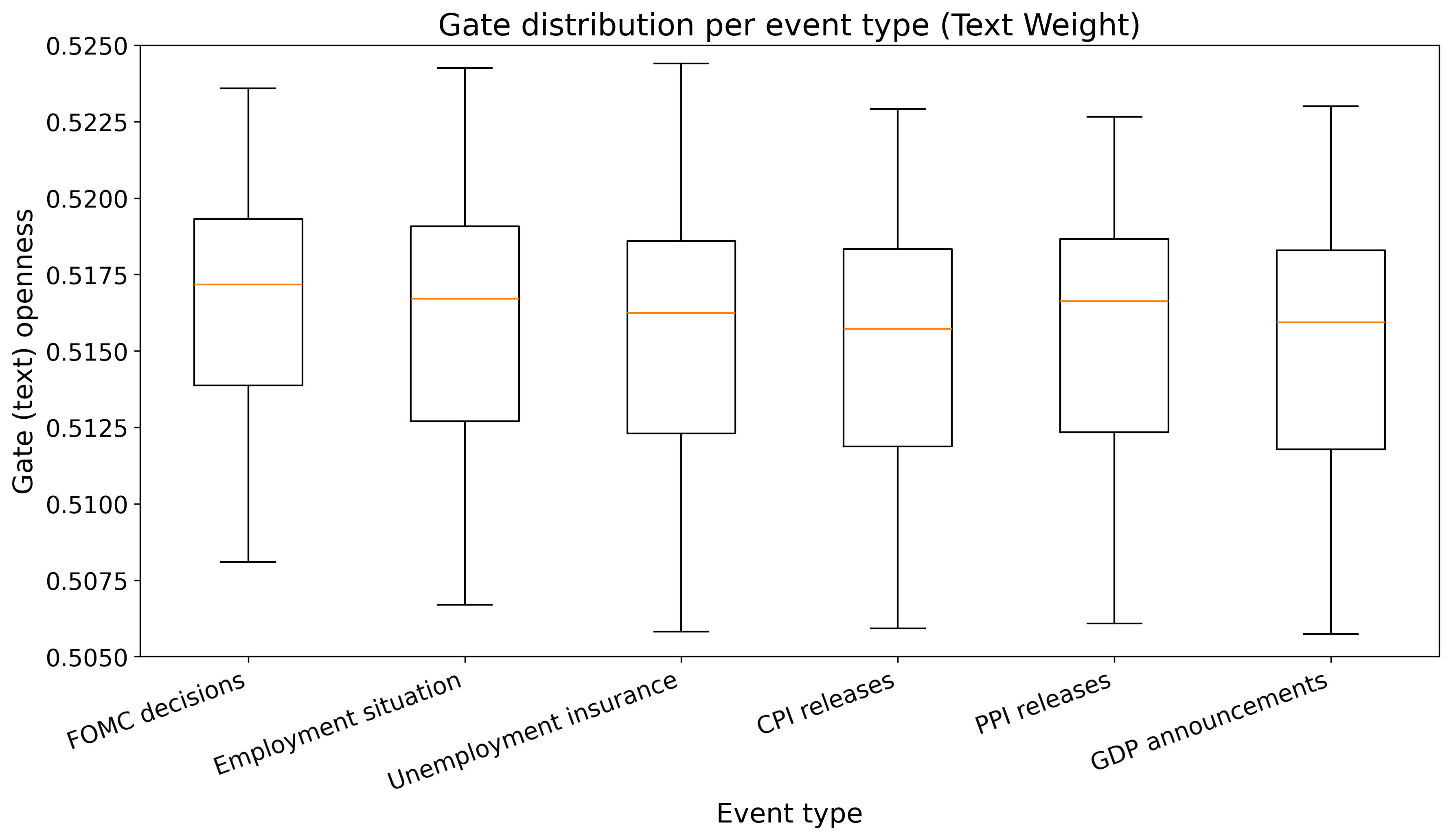}
        \caption{\textsc{S\&P500}: gate openness by event type.}
        \label{fig:gate_sp500_events}
    \end{subfigure}

    \vspace{0.5em}

    \begin{subfigure}[t]{0.48\textwidth}
        \centering
        \includegraphics[width=\textwidth]{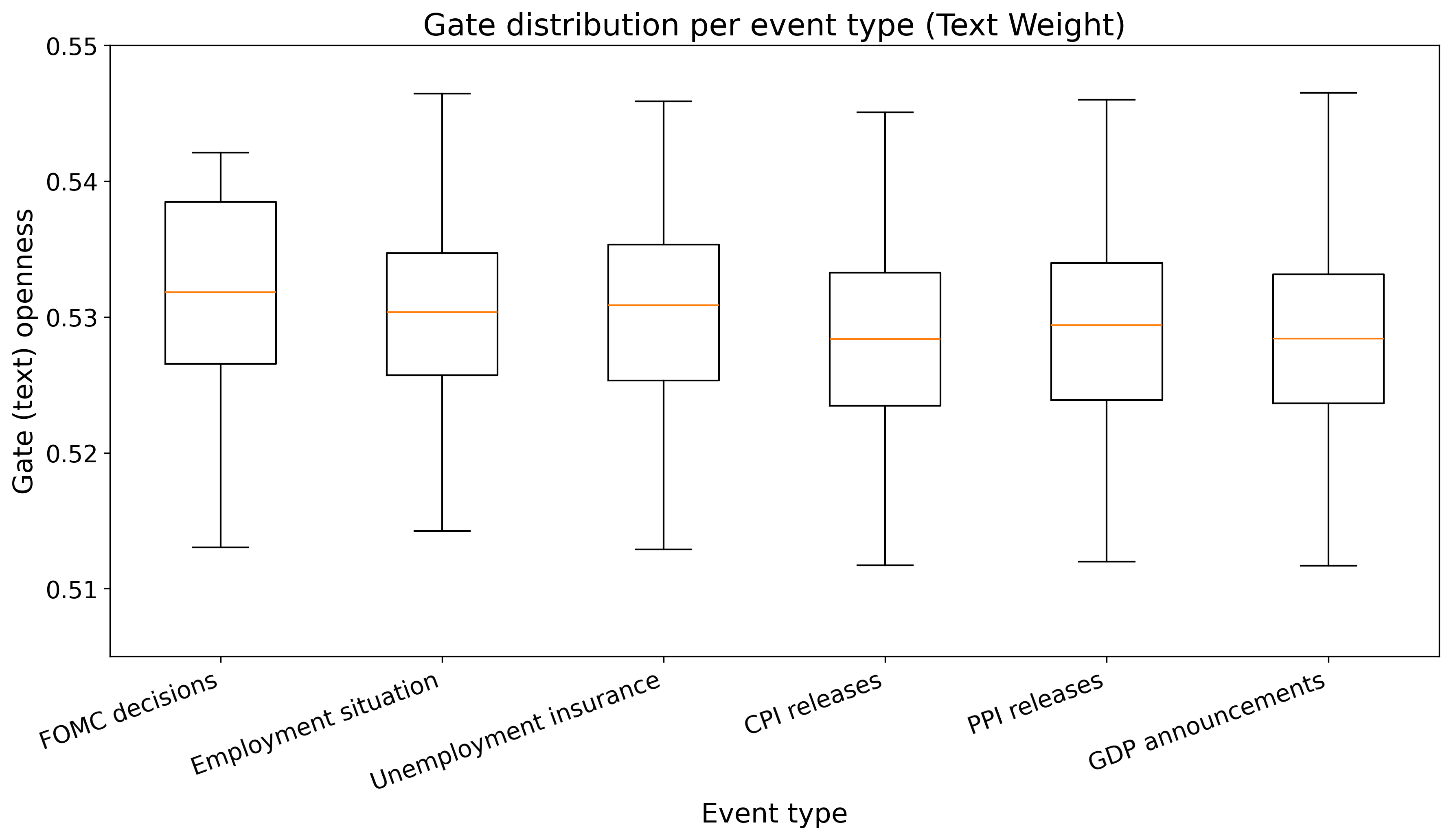}
        \caption{\textsc{INDU}: gate openness by event type.}
        \label{fig:gate_indu_events}
    \end{subfigure}
    \hfill
    \begin{subfigure}[t]{0.48\textwidth}
        \centering
        \includegraphics[width=\textwidth]{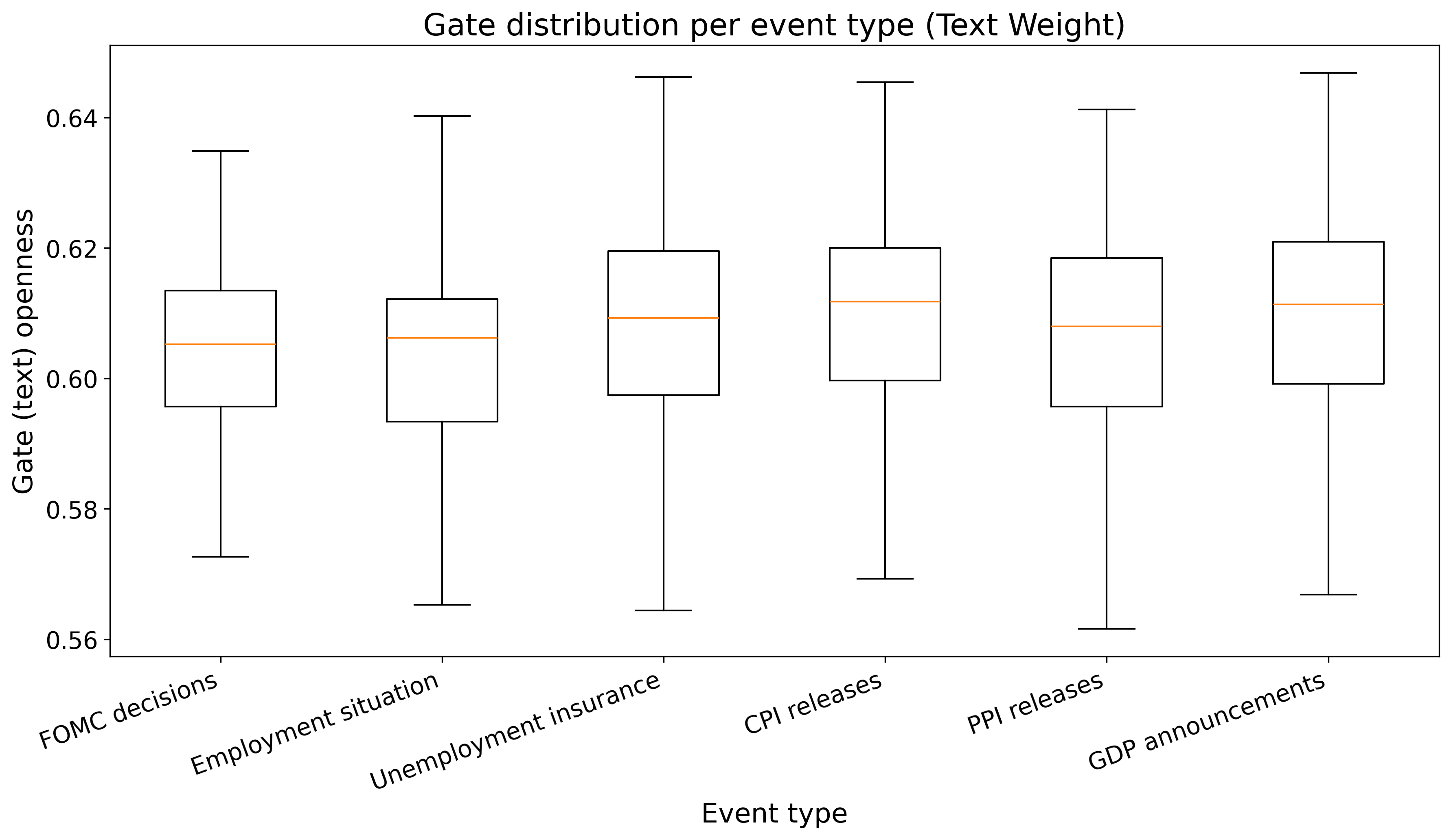}
        \caption{\textsc{USGG1M}: gate openness by event type.}
        \label{fig:gate_usgg1m_events}
    \end{subfigure}

    \caption{Distributions of learned text-gate openness.
    (a) Aggregate gate openness across datasets.
    (b)--(d) Gate openness stratified by macroeconomic event type for \textsc{S\&P500}, \textsc{INDU}, and \textsc{USGG1M}.}
    \label{fig:gate_appendix}
\end{figure*}

\section{Additional Analysis of Gate Openness}
\label{app:gate_analysis}

To better understand how GS-Fuse uses event text, we examine the distribution of the learned text-gate openness across datasets and event types. As shown in Fig.~\ref{fig:gate_appendix}, gate values for the two equity indices (\textsc{S\&P500}, \textsc{INDU}) concentrate around $0.51$--$0.53$, whereas \textsc{USGG1M} exhibits a substantially higher central tendency (around $0.60$). This indicates that GS-Fuse relies more heavily on textual information when forecasting short-term interest rates than when forecasting equity indices, consistent with the stronger sensitivity of bond markets to macroeconomic announcements. 

Within each dataset, the gate distributions over different macroeconomic event types (FOMC decisions, employment situation, unemployment insurance, CPI releases, PPI releases, and GDP announcements) vary only within a narrow band, suggesting broadly consistent usage of text across heterogeneous announcements. Within this band, monetary policy and inflation-related events (FOMC, CPI, PPI) tend to show slightly higher openness than labor-market releases, indicating that the model assigns somewhat greater incremental predictive value to events that directly convey policy stance and price-level information.

\vspace{-1em}

\section{Case Study: Token Alignment Analysis}

We randomly sample two held-out event scripts, one \textbf{FOMC decision} and one \textbf{PPI release}, to qualitatively inspect whether the \emph{top aligned} (highlighted) tokens within each script are economically meaningful rather than boilerplate. The following subsections illustrate detailed analyses.

\vspace{-1.1em}

\subsection{Case 1: FOMC Decision Script}
\newtcbox{\token}{on line, 
  arc=2pt, colback=blue!10, colframe=blue!30, 
  before upper=\rule[-3pt]{0pt}{10pt}, boxrule=0.5pt, 
  boxsep=0pt, left=2pt, right=2pt, top=1pt, bottom=1pt,
  fontupper=\small\ttfamily}

\begin{tcolorbox}[colback=white, colframe=gray!50, arc=0pt, outer arc=0pt, title=]
At the \token{meeting}, the \token{committee} decided to keep the \token{federal} \token{funds} \token{rate} \token{target} \token{range} at \token{0-0.25\%}. The \token{committee} also decided to end \token{its} asset purchase program in October and \token{maintain} \token{its} existing \token{policy} of reinvesting principal payments. The \token{committee} noted that economic activity was expanding at a moderate pace, with \token{labor} \token{market} \token{conditions} improving further, and \token{inflation} running below \token{its} \token{2\%} \token{longer-run} goal.
\end{tcolorbox}

\vspace{-1em}

From the highlighted tokens above, we observe that the model consistently attends to the policy instrument and its numerical setting, aligning the phrase ``federal funds rate target range'' with the corresponding interval ``0–0.25\%.'' At the same time, it emphasizes mandate-relevant clauses such as ``labor market conditions'' and ``inflation… 2\% longer-run,'' which govern expectations about future policy. These patterns suggest that the token-wise alignment module is sensitive to both the current policy threshold and the macroeconomic conditions that define the event's market impact.

\vspace{-1em}

\subsection{Case 2: PPI Report Event Script}

\begin{tcolorbox}[
  colback=green!2!white, 
  colframe=gray!80, 
  arc=0pt, 
  outer arc=0pt, 
  boxrule=0.5pt, 
  leftrule=3pt, 
  fonttitle=\small\sffamily\bfseries,
  coltitle=black,
  colbacktitle=gray!15
]
\small
The \token{PPI} for \token{food} and \token{beverages} \token{increased} by \token{1.3\%} from May \token{2021} to June \token{2021}... 
The \token{PPI} for \token{transportation} \token{services} \token{increased} by \token{7.2\%}, driven by higher prices for gasoline, diesel fuel, and air \token{travel}. 
The \token{PPI} for \token{services} \token{related} to \token{transportation} \token{activities} \token{increased} by \token{23.5\%}.
The \token{PPI} for \token{investment} \token{services} \token{increased} by \token{25.7\%}. 
The \token{PPI} for \token{real} \token{estate} \token{services} \token{increased} by \token{5.5\%}. 
The \token{PPI} for \token{rental} and \token{leasing} of \token{goods} \token{increased} by \token{17.8\%}.
\end{tcolorbox}

From this case, we note that the alignment module consistently selects the index name (``PPI''), the sector descriptors (``food and beverages,'' ``transportation services,'' ``investment services,'' ``real estate services,'' ``rental and leasing of goods''), and the associated directional and quantitative cues (``increased,'' along with the percentage changes such as ``1.3\%,'' ``7.2\%,'' ``23.5\%,'' ``25.7\%,'' ``5.5\%,'' ``17.8\%''). This pattern suggests that the token-alignment mechanism is sensitive not only to the overall index label but also to the decomposition of inflation across sectors and the relative size of each sector's price change, which are exactly the components that matter for downstream market reactions.

\vspace{-1em}

\section*{Limitations and Future Directions}

While GS-Fuse delivers consistent gains over strong unimodal and multimodal baselines across five major U.S. assets and three horizons, our study still has a bounded scope. We focus on macroeconomic announcements in the U.S. and a small set of highly liquid benchmarks at 5-minute frequency, and we evaluate point forecasts (MSE/MAE) of prices and yields rather than volatility or tail risk. Extending GS-Fuse to additional asset classes, geographies, frequencies, and risk-sensitive or strategy-level evaluations is a natural direction for future work rather than a limitation of the framework itself.

From a modeling perspective, we instantiate GS-Fuse with compact LLM backbones and two representative time-series foundation models, and keep these encoders mostly frozen to respect realistic GPU budgets. We adopt commonly used hyperparameters for baselines and set GS-Fuse–specific ones by conceptual choices, so further tuning could in principle improve performance. Our fusion module also uses a smooth soft gate for stability; sharper gating settings could be explored in future work.

Finally, our Granger-style supervision should be viewed as a principled predictive utility signal rather than a full structural causal claim. We assume macro texts are exogenous information shocks and do not explicitly model all possible confounders or regime shifts. In practice, deploying GS-Fuse in high-stakes settings would benefit from complementary components such as regime detection, uncertainty quantification, and more sophisticated document compression or longer-context encoders, which we leave for future work.

\end{document}